\documentclass{article}

\usepackage[nonatbib, final]{neurips_2021}

\usepackage[square,numbers]{natbib}
\usepackage[utf8]{inputenc} 
\usepackage[T1]{fontenc}    
\usepackage{hyperref}       
\usepackage{url}            
\usepackage{booktabs}       
\usepackage{amsfonts}       
\usepackage{nicefrac}       
\usepackage{microtype}      
\usepackage{graphicx}
\usepackage{url}            
\usepackage{booktabs}       
\usepackage{amsfonts}       
\usepackage{nicefrac}       
\usepackage{xcolor}
\usepackage{hyperref}
\usepackage{multirow}
\usepackage{xspace}
\usepackage{amsmath}
\usepackage{mathtools}
\usepackage{listings}
\usepackage{amsfonts}
\usepackage{amssymb,algorithm}
\usepackage{algpseudocode}
\usepackage{textcomp}
\usepackage{enumitem}
\usepackage{cleveref}
\usepackage{subcaption}
\usepackage{float}
\usepackage{wrapfig}
\usepackage{enumitem}
\newcommand{\llc}{\ensuremath{{\rm LLC}}\xspace}

\crefformat{section}{\S#2#1#3}
\crefformat{subsection}{\S#2#1#3}

\hypersetup{
    colorlinks=true,
    linkcolor=blue,
    filecolor=magenta,      
    urlcolor=cyan,
}
\begin{document}

\title{\llc: Accurate, Multi-purpose \textbf{L}earnt \textbf{L}ow-dimensional Binary \textbf{C}odes}
\author{
Aditya Kusupati$^\dagger$,\\ \textbf{\hspace{-4mm}Matthew Wallingford$^\dagger$, Vivek Ramanujan$^\dagger$, Raghav Somani$^\dagger$, Jae Sung Park$^\dagger$, Krishna Pillutla$^\dagger$,}\\ \textbf{Prateek Jain$^\ddagger$, Sham Kakade$^\dagger$ and Ali Farhadi$^\dagger$}\\
$^\dagger$University of Washington, 
$^\ddagger$Google Research India\\
\texttt{\hspace{-4mm}\{kusupati,mcw244,ramanv,raghavs,jspark96,pillutla,sham,ali\}@cs.washington.edu},\\ \texttt{prajain@google.com}
}

\maketitle
\begin{abstract}

Learning binary representations of instances and classes is a classical problem with several high potential applications. In modern settings, the compression of high-dimensional neural representations to low-dimensional binary codes is a challenging task and often require large bit-codes to be accurate. In this work, we propose a novel method for \textbf{L}earning \textbf{L}ow-dimensional binary \textbf{C}odes (\llc) for instances as well as classes. Our method does {\em not} require any side-information, like annotated attributes or label meta-data, and learns extremely low-dimensional binary codes ($\approx 20$ bits for ImageNet-1K). The learnt codes are super-efficient while still ensuring {\em nearly optimal} classification accuracy for ResNet50 on ImageNet-1K. We demonstrate that the learnt codes capture intrinsically important features in the data, by discovering an intuitive taxonomy over classes. We further quantitatively measure the quality of our codes by applying it to the efficient image retrieval as well as out-of-distribution (OOD) detection problems. For ImageNet-100 retrieval problem, our learnt binary codes outperform $16$ bit HashNet using only $10$ bits and also are as accurate as $10$ dimensional real representations. Finally, our learnt binary codes can perform OOD detection, out-of-the-box, as accurately as a baseline that needs $\approx3000$ samples to tune its threshold, while we require {\em none}. Code is open-sourced at \url{https://github.com/RAIVNLab/LLC}.

\end{abstract}
\section{Introduction}
\label{sec:intro}

Embedding data in low-dimensional binary space is a long-standing machine learning problem~\cite{wang2015learning}. The problem has received a lot of interest in the computer vision (CV) domain, where the goal is to find binary codes that capture the key semantics of the image, like, objects present in the image or interpretable attributes. Section~\ref{sec:rw} covers the literature on learning binary codes and their applications. 

In addition to learning semantically meaningful representations of the instances, low-dimensional binary codes allow efficiency in a variety of large-scale machine learning (ML) applications. Low-dimensional codes are crucial in extreme classification with millions of classes~\citep{bhatia2015sparse, jain2019slice, dahiya2021deepxml} and also critical in efficient large-scale retrieval settings~\citep{liu2016deep,datar2004locality,weyand2020google}. 

Compressing information into binary codes is challenging due to its highly non-smooth nature while requiring the preservation of relevant information in an instance/class. This might explain the lack of good classification accuracy for existing classification-based embedding techniques~\cite{hsu2009multi,cisse2013robust}. To address that, traditional methods often relied on side-information like attributes to construct class codes and then use that to learn the instance codes~\citep{deng2011hierarchical,akata2015label}. 

Learning binary embeddings can be posed in a variety of formulations like pairwise optimization~\cite{kulis2009fast} or unsupervised learning~\citep{carreira2015hashing,salakhutdinov2007learning}, in this work we focus on learning binary codes using a given labeled multi-class dataset, e.g., ImageNet-1K. This allows us to couple the representation (code) learning of both {\em instances} and  {\em classes} thus enabling us to capture the underlying semantic structure efficiently to assist in downstream tasks like classification, retrieval etc. 

We propose \llc, a method to learn {\em both} class and instance codes via the standard classification task and its setup {\em without any side-information}. Our Learning Low-dimensional binary Codes (\llc) technique, formulates the embedding (code) learning problem as that of learning a low-dimensional binary embedding of a standard deep neural ``backbone''. Instead of directly training for the low-dimensional binary instance codes, we propose a two-phase approach. In the first phase, \llc learns low-dimensional ($k$-bit) binary codes for classes that capture semantic information through a surrogate classification task. Then in the second phase, \llc uses these learnt class codes as an efficient alternative to learning instance codes in sub-linear cost (in the number of classes, $L$) using the Error-Correcting Output Codes (ECOC) approach~\citep{dietterich1994solving}. This two-phase pipeline helps in the effective distillation of required semantic similarity between instances through the learnt class codes. For example, on ImageNet-1K with ResNet50, \llc is able to learn tight $20$-bit codes that can be used for {\em efficient classification} and achieve $74.5\%$ accuracy compared to the standard baseline $77\%$ on ImageNet-1K (Section~\ref{sec:classification_expts}). Furthermore, we observe that the learnt $20$-bit class codes capture intuitive taxonomy over classes (Figure~\ref{fig:heir}) while the instance codes retain the distilled class similarity information useful in efficient retrieval and OOD detection. 

{\bf Retrieval.} To further study, the effectiveness of our learnt binary codes, we apply them to hashing-based efficient retrieval, where the goal is to retrieve a large number of similar instances with the same class label in top retrieved samples. Deep supervised hashing is a widely studied problem with several recent results~\cite{cao2017hashnet,su2018greedy} which are designed {\em specifically} for the learnt hashing-based retrieval. Interestingly, our learnt instance codes through the \llc routine provide strikingly better performance while not being learnt explicitly for hashing. For eg., using AlexNet, with just $32$-bit codes we are can provide $5.4\%$ more accurate retrieval than HashNet's $64$-bit codes on ImageNet-100 (Section~\ref{sec:retrieval_expts}). 

{\bf OOD Detection.} We similarly apply \llc based learnt binary codes to detect OOD instances~\citep{hendrycks2016baseline}. We adopt a simple approach based on our binary codes: if an instance is not within a Hamming distance of $1$ to any class codes, we classify it as OOD. That is, we do not fine-tune our OOD detector for the new domain, which is critical in practical settings. In contrast, baseline techniques for OOD detection require a few samples (eg., $\approx3000$ for ImageNet-750) from the OOD domain to fine-tune thresholds, while we require {\em no} samples yet reaching similar OOD detection (Section~\ref{sec:ood_expts}). 

In this work, we make the following key contributions:
\begin{itemize}[leftmargin=*]\vspace{-1mm}
    \itemsep 0pt
    \topsep 0pt
    \parskip 2pt
    \item \llc method to learn semantically-rich low-dimensional binary codes for both classes \& instances.
    \item Show that the learnt codes enable accurate \& efficient classification: ImageNet-1K with $20$-bits.
    \item Apply \llc to image retrieval task, and demonstrate that it comfortably outperforms the instance code learning methods for hashing-based retrieval on ImageNet-100.
    \item Finally, use codes from \llc for strong \& sample efficient OOD detection in practical settings. 
\end{itemize}

\section{Related Work}
\label{sec:rw}
Binary class codes were originally aimed at sub-linear training and prediction for multi-class classification. The Error-Correcting Output Codes (ECOC) framework~\citep{dietterich1994solving,allwein2000reducing,escalera2010error} reformulated multi-class classification as multi-label classification using $k$-bit codes per class (codebook). The learning of optimal codebook is NP-complete~\citep{crammer2001algorithmic} which lead to use of random codebooks~\citep{hsu2009multi,cisse2013robust} in traditional ML. However, there were a few codebook learning~\citep{bengio2010label,zhang2011multi,weston2011wsabie,bautista2016learning} and construction schemes using side-information from other modalities~\citep{akata2015label}. The lack of a strong learnable feature extractor often deterred the gains these codebooks provide for the classification and effective learning of instance codes. Attribute annotations can also help in constructing class codes~\citep{akata2015evaluation}. These binary codes are either explicitly annotated~\citep{farhadi2009describing} or discovered~\citep{rastegari2012attribute,ferrari2007learning}. Attributed-based learning also ties into leveraging the class codes for zero/few-shot learning~\citep{lampert2009learning,lampert2013attribute,akata2015label,norouzi2013zero} expecting some form of interpretability.

Most methods that use class codes as supervision can produce instance codes~\citep{dietterich1994solving}. However, the standalone literature of instance codes comes from requirements in large-scale application like retrieval (hashing). In the past, most hashing techniques that created instance codes were based on random projections~\citep{datar2004locality,cisse2013robust,choromanska2016binary}, semantics~\citep{salakhutdinov2009semantic,deng2011hierarchical} or learnt through metric learning~\citep{kulis2009fast,kulis2009kernelized,norouzi2012hamming,kulis2009learning}, clustering~\citep{weiss2008spectral,salakhutdinov2007learning} and quantization~\citep{gong2012iterative}. Deep learning further helped in learning more accurate hashing functions to generate instance codes either in an unsupervised~\citep{carreira2015hashing,shen2018unsupervised} or supervised~\citep{liu2016deep,cao2017hashnet,su2018greedy,yuan2018relaxation} fashion. We refer to~\citep{luo2020survey,zhang2020survey,wang2015learning} for a more thorough review on deep hashing methods. 

Finally, embedding-based classification~\citep{cisse2013robust,yu2014large,bhatia2015sparse,Guo2019BreakingTG} enables joint low-dimensional representation learning for both classes and instances with an eye on sub-linear training and prediction costs. After distilling the key ideas from the literature, we aim to a) learn semantically rich low-dimensional representations for both classes and instances together, b) have these representations in the binary space, and c) do this with minimal dependence on side-information or metadata.

\llc, to the best of our knowledge - for the first time, jointly learns low-dimensional binary codes for both classes and instances using a surrogate classification task, without any side-information (Section~\ref{sec:llc}). The learnt class codes capture intrinsic information at the semantic level that helps in discovering an intuitive taxonomy over classes (Figure~\ref{fig:heir}). The learnt class codes then anchor the instance code learning which results in tight and accurate low-dimensional instance codes further used in retrieval (Section~\ref{sec:retrieval_expts}). Finally, both the learnt class and instance codes power extremely efficient yet accurate classification (Section~\ref{sec:classification_expts}) and out-of-distribution detection (Section~\ref{sec:ood_expts}).
\section{Learning Low-dimensional Binary Codes}
\label{sec:llc}
The goal is to learn a binary embedding (code) function $g\colon{\mathcal{X}} \to \{-1,1\}^k$ where $\mathcal{X}$ is the input domain and $k$ is the dimensionality of the code. We focus on learning embeddings using a labelled multi-class data~\citep{hsu2009multi}. That is, suppose we are given a labelled dataset $\mathcal{D}=\{(x_1, y_1), \ldots, (x_n,y_n)\}$ where $x_i\in \mathcal{X}$ is an input point and $y_i \in [L]$ is the label of $x_i$ for all $i\in[n]$. Then, the goal is to learn an instance embedding function $g\colon {\cal X}\to \{+1,-1\}^k$ {\em and} class embeddings $h_q\in\{+1,-1\}^k$ for all $q\in[L]$ such that $g(x_i) = h_{y_i}$ and $g(x_i) = g(x_j)$ if and only if $y_i=y_j$. 

Intuitively, for large-scale datasets,  $g(x)$ and $h_q$ should capture key semantic information to provide accurate classification, thus allowing their use in application domains like retrieval or OOD detection. Note that while we present our technique for learning embeddings using multi-class datasets, it applies more generally to multi-labeled datasets as well.

\paragraph{Instance and Class Code Parameterization.} For learning such embedding function, we assume access to a deep neural architecture $F(\, \cdot\,  ;\theta_F)\colon \mathcal{X} \rightarrow \mathbb{R}^d$ that maps the input $x\in {\cal X}$ to a $d$-dimensional real-valued representation. $\theta_F$ is a learnable parameterization of the network; we drop $\theta_F$ from $F$ wherever the meaning is clear from the context. For example, ResNet50 is one such network that encodes $224\times 224$ RGB images into $d=2048$ dimensions. 

Now, given a network $F$ and $x\in {\cal X}$, we formulate embedding function of $x$ and the corresponding multiclass prediction scores $\hat{y}\in \mathbb{Z}^L$ as:
\begin{equation}
    \label{eq:embedx}
    g(x) \coloneqq B\left(\mathbf{P}\cdot F(x; \theta_F)\right), \quad \hat{y} \coloneqq B(\mathbf{C}) \cdot g(x)\ ,
\end{equation}
where $\mathbf{P}\in \mathbb{R}^{k\times d}$ maps $F(x)$ into $k$-dimensions and $B(a)=\mathrm{sign}(a)\in\{+1, -1\}$ is the standard binarization/sign operator applied elementwise (with the assumption $\mathrm{sign}(0)=+1$). Finally, $\mathbf{C} \in \mathbb{R}^{L\times k}$, and $\hat{y} = B({\mathbf{C}})\cdot g(x)$ represents the scores of each class for an input $x$. Note that for a class $\ell\in[L]$, $B(\mathbf{C}_\ell)$ (where $\mathbf{C}_\ell$ represents the $\ell$-th row of $\mathbf{C}$) is the learnt binary class embedding (code) of class $\ell\in [L]$, and $g(x)=B(\mathbf{P}\cdot F(x; \theta_F))$ is the learnt instance embedding (code) of instance $x$. Note that~\eqref{eq:embedx} is a general purpose formulation for the problem of learning class and instance codes.

\subsection{The \llc Method}

\paragraph{Phase 1: \em{Codebook Learning} -- $B(\mathbf{C})$.} Given labelled examples ${\cal D}$, we use standard empirical risk minimization to learn a multi-class classifier, i.e., we solve
\begin{equation}
    \label{eq:phase1}
    \min_{\mathbf{C}, \mathbf{P}, \theta_F} \sum_{(x_i,y_i)\in {\cal D}} {\cal L}\left(B(\mathbf{C}) \cdot (\mathbf{P}\cdot F(x_i; \theta_F))\ ;\ y_i\right) \ ,
\end{equation}
where ${\cal L}\colon \mathbb{R}^L\times [L] \to \mathbb{R}_+$ is the standard multi-class softmax cross-entropy loss function. This is a standard optimization problem that can be solved using standard gradient descent methods or other sub-gradient based optimizers. However, one challenge is that $B(\mathbf{C})$ is a binary matrix and $B$ is a binary function, so the gradients are $0$ almost everywhere. Instead, we use the Straight-Through Estimator (STE)~\citep{bengio2013estimating} technique popular in binary neural networks domain~\citep{rastegari2016xnor}, to optimize for $\mathbf{C}$ through the binarization. Intuitively, STE uses binarization/sign function in the forward pass, but in the backpropagation phase, it allows the gradients to flow straight-through as if it were real-valued. The codebook, $B(\mathbf{C})$ refers to the collection of $k$-bit class codes learnt in this process.
 
For ImageNet-1K, we learnt unique binary codes, $B(\mathbf{C}_\ell)$, for every class $\ell\in[L]$ of the 1000 classes using only $20$-bits, only twice the information-theoretic limit. As with the class representations from a linear classifier, these class codes do capture intrinsically important features that help in discovering intuitive taxonomy over classes (Section~\ref{sec:hierarchy}) among various applications (Section~\ref{sec:apps+exps}).

\paragraph{Phase 2: \em{Instance Code Learning} -- $B(\mathbf{P}\cdot F(x; \theta_F))$.} Several existing techniques model $\mathbf{C}$ and $\mathbf{P}$ in different ways to learn an embedding function similar to~\eqref{eq:embedx}. However, these methods often try to only learn instance codes and have challenges in maintaining high accuracy~\citep{cao2017hashnet,carreira2015hashing} in a variety of applications because optimization problem (\ref{eq:phase1}) is challenging and might lead to significantly sub-optimal classification error. For example, for ImageNet-1K classification with ResNet50, the accuracy for our trained model (20-bits) at this stage is $72.5\%$ compared to the standard $77\%$.

To remedy this, we further optimize our embeddings using the ECOC framework~\citep{dietterich1994solving} for multi-class classification, which essentially transforms the multi-class problem into a multi-label problem, which in turn is $k$ independent binary classification problems. That is, we use the $k$-bit codes learnt for each class as the supervision to further train $F(\, \cdot\, ; \theta_F)$ and $\mathbf{P}$: 
\begin{equation}\label{eq:phase2}
\min_{\theta_F, P} \sum_{(x_i,y_i)\in{\cal D}} \sum_{j=1}^k \textrm{BCE}\left(\sigma(\mathbf{P}_j\cdot F(x_i; \theta_F))\ ;\  \left(B(\mathbf{C}_{y_i,j})+1\right)/2\right),
\end{equation}
where $\sigma$ is the sigmoid/logistic function, $\textrm{BCE}$ is the binary cross-entropy loss between the $j$-th bit of instance $x_i$'s embedding, and the $j$-th bit extracted from the class embedding of it's label $y_i$ (the function $z\mapsto (z+1)/2$ is used to map $\{+1,-1\}$ to $\{1,0\}$ to make it a simple binary classification problem per each bit). We use gradient based optimization to learn $\theta_F$ and $\mathbf{P}$. As mentioned earlier, ECOC framework allows us to correct errors in classification. For example, with just $20$ bits on ImageNet-1K dataset, the method now achieves $74.5\%$ accuracy with ResNet50 backbone.

The advantage of this two-phase pipeline where we start with a codebook learning for classes is that the cost of learning instances codes reduces to a bottleneck of $k$-dims ($\ll L$) instead of the usual $L$ . Furthermore, these learnt low-dimensional binary codes for both classes and instances help in large-scale applications via efficient classification and retrieval (see Section~\ref{sec:apps+exps}). Note that, unlike attribute-based methods~\citep{lampert2013attribute}, we do {\em not} require additional meta-data, but learn binary codes by only using the standard classification task. This also circumvents the potential instabilities of pairwise optimization in instance binary code learning which often leads to poor class codes due to codebook collapse. At the end of \llc routine, we have learnt the instance codes, $B(\mathbf{P}\cdot F(x; \theta_F))$, and class codes, $B(\mathbf{C})$ to be used for downstream applications. Algorithm~\ref{alg:llc} presents \llc in full.

\begin{algorithm}[b!]
\caption{The \llc Method}
\label{alg:llc}
\begin{algorithmic}[1]
\Require ${\cal D}$, $F$ and $B$
\Ensure $\mathbf{C}$, $\mathbf{P}$ and $\theta_F$
\State {\bf Codebook Learning -- $B(\mathbf{C})$}: Solve~\eqref{eq:phase1} using ERM and STE to get $\mathbf{C}$, $\mathbf{P}$ and $\theta_F$ -
\[
    \mathbf{C}, \mathbf{P}, \theta_F \gets \mathop{\arg\min}_{\mathbf{C}, \mathbf{P}, \theta_F} \sum_{(x_i,y_i)\in {\cal D}} {\cal L}\left(B(\mathbf{C}) \cdot (\mathbf{P}\cdot F(x_i; \theta_F))\ ;\ y_i\right)\ .
\]
\State {\bf Instance Code Learning -- $B(\mathbf{P}\cdot F(x; \theta_f))$}: Further optimize $\mathbf{P}$ and $\theta_F$ by solving~\eqref{eq:phase2} using ECOC framework and ERM by fixing $\mathbf{C}$ -
\[
    \theta_F, P \gets \mathop{\arg\min}_{\theta_F, P} \sum_{(x_i,y_i)\in{\cal D}} \sum_{j=1}^k \textrm{BCE}\left(\sigma(\mathbf{P}_j\cdot F(x_i; \theta_F))\ ;\  \left(B(\mathbf{C}_{y_i,j})+1\right)/2\right)\ .
\]
\end{algorithmic}
\end{algorithm}

Overall, we present a simple yet scalable method to learn low-dimensional (exact) binary codes for both classes and instances which in turn could power multi-class classification with sub-linear costs (in terms of $L$) and efficient retrieval for large-scale applications. Using our method, we can consistently learn unique low-dimensional binary codes for all 1000 classes in ImageNet-1K using only $20$-bits (which is twice the information-theoretic limit of $\lceil\log_2 (1000)\rceil$). Next, we discuss the learnt codebook's intrinsic information about the classes and their structure.

\subsection{Discovered Taxonomy and Visualizations}
\label{sec:hierarchy}
\begin{figure}[ht!]
\centering
\resizebox{\columnwidth}{!}{%
\includegraphics[width=\columnwidth, angle =-90 ]{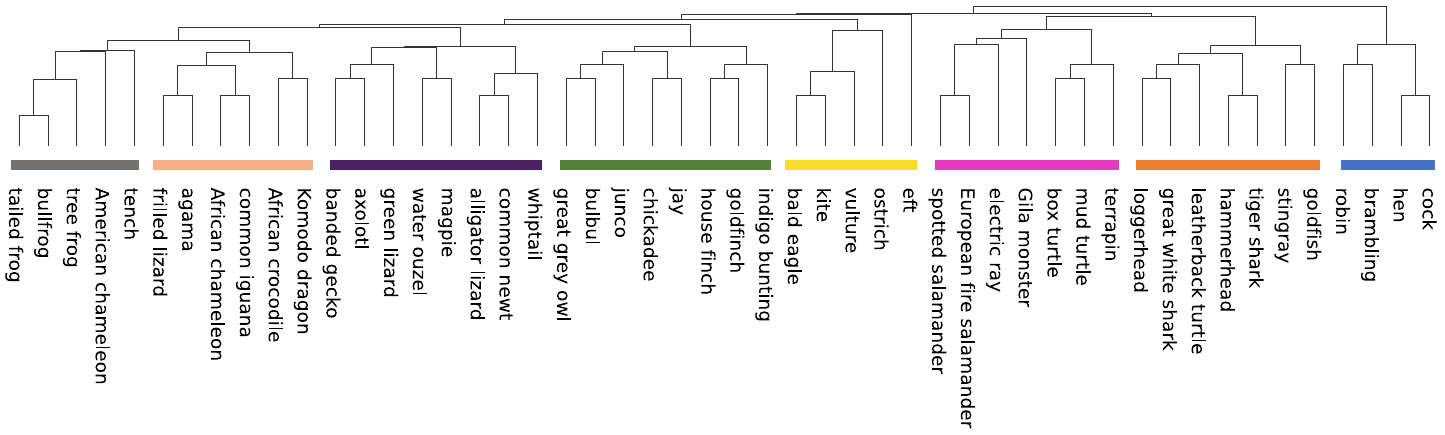}
}
\caption{\small Discovered taxonomy over 50 classes of ImageNet-1K using the learnt 20-bit class codes. Related species are well clustered while pushing away unrelated ones. Figure~\ref{fig:heir_code} in Appendix~\ref{sec:quant_hier} contains the codebook.}
\label{fig:heir}
\end{figure}
After learning the $20$-bit binary codebook for 1000 classes of ImageNet-1K, we used the class representation from $B(\mathbf{C})$ of the first 50 classes to discover an intuitive taxonomy through agglomerative clustering~\citep{murtagh2012algorithms}. Figure~\ref{fig:heir} shows the discovered hierarchy. This hierarchy effectively separates birds from amphibians; frogs and chickens are on extremes of the taxonomy and brings species with shared similarities closer (lizards \& crocodiles; marine life). While the taxonomy is not perfect, the $20$-bits do capture enough important information that can be used downstream.

\begin{figure}[ht!]
\centering
\begin{tabular}{cc}
  \includegraphics[width=.5\textwidth]{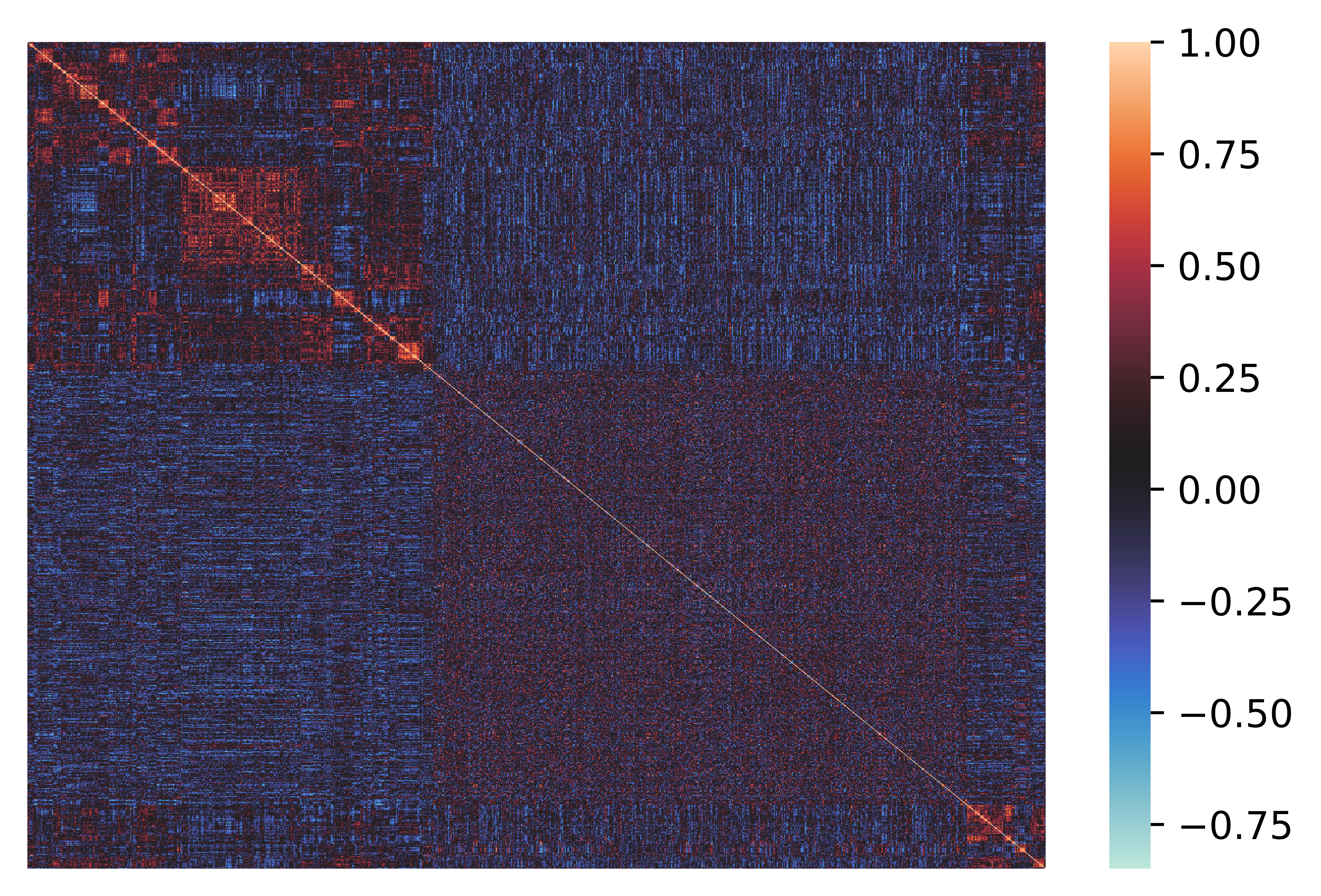}&
  \includegraphics[width=.5\textwidth]{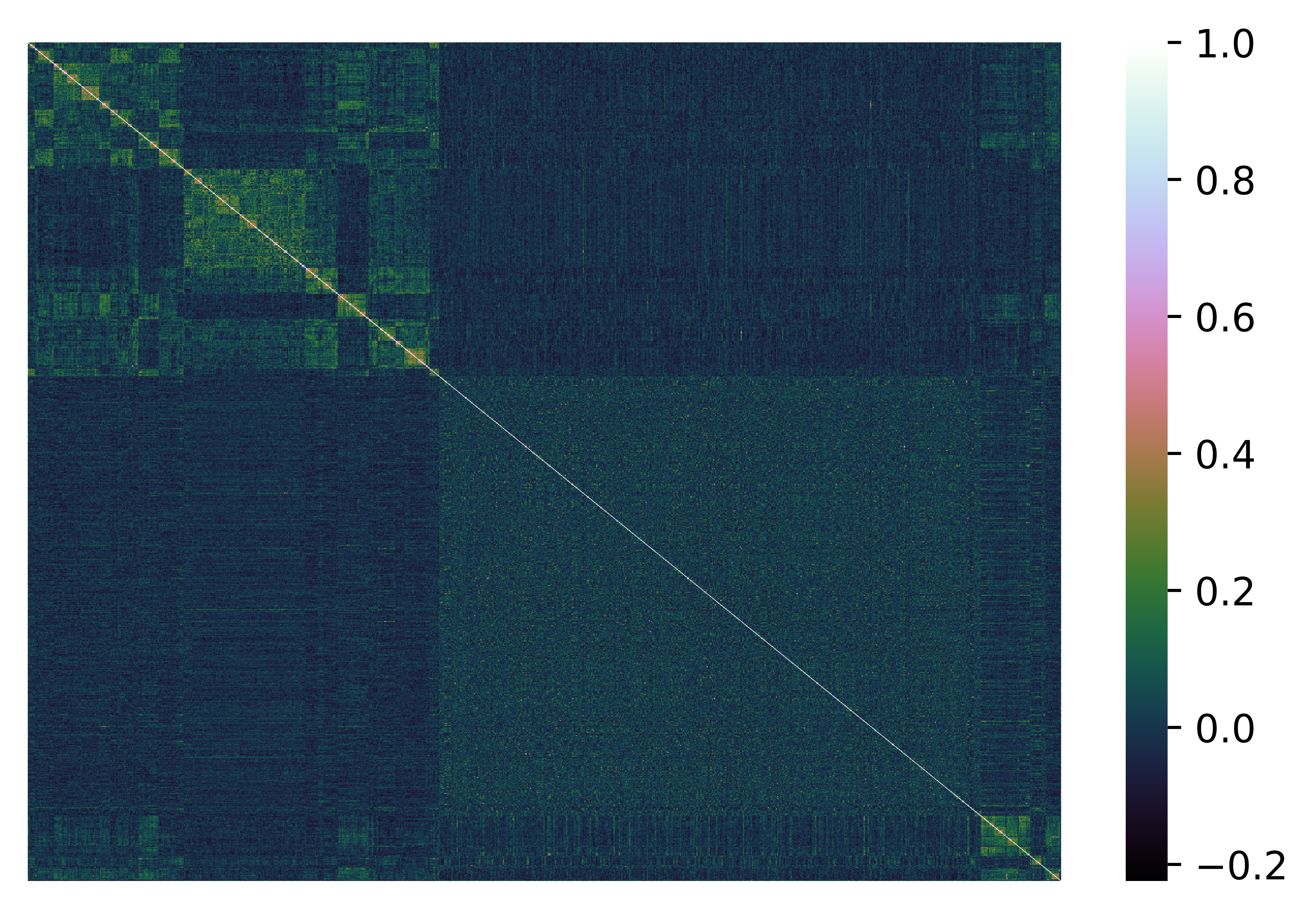}\\
(a) 20-bit codes&(b) 2048-d real representations
\end{tabular}
\caption{\small The pair-wise inner product heat maps of class representations a) learnt $20$-bit codes \& b) learnt $2048$ dimensional real representations for the 1000 classes in ImageNet-1K. Similar sub structures are highlighted in both heatmaps and often correspond to local hierarchy present in the classes thus making a case that 20-bit codes distill enough information to capture hierarchy of the classes.}
\label{fig:heat}
\vspace{-2mm}
\end{figure}

Figure~\ref{fig:heat} shows the pair-wise inner-product heat maps for all the 1000 classes using $20$-bits and $2048$-dimensional real representation; the comparison reveals that $20$-bits indeed highlights the same substructures as the higher dimensional real-valued embeddings. Appendix~\ref{sec:quant_hier} has a more detailed discussion about quantitatively evaluating the discovered hierarchy and more visualizations.

\section{Applications}
\label{sec:apps+exps}
In this section, we discuss three applications of the learnt low-dimensional binary codes: 1) efficient multi-class classification (Section~\ref{sec:classification_expts}), 2) efficient retrieval (Section~\ref{sec:retrieval_expts}), and 3) out-of-the-box out-of-distribution (OOD) detection  (Section~\ref{sec:ood_expts}).  We also present ablation studies on codebook learning, feature separability and classification (Section~\ref{sec:ablations}).

\subsection{Efficient Multi-class Classification}
\label{sec:classification_expts}
Recall that the proposed \llc algorithm outputs a) the learnt class codes (codebook), $B(\mathbf{C})$ and b) an  {\em encoder} that produces instance codes, $B(\mathbf{P}\cdot F(\mathbf{x};\theta_F))$ for $x$. We define a class codebook as a collection of $L$ binary vectors, one for each class in the dataset, that can then be used for classification of a test instance $x$. We can use several ``decoding" routines to classify an instance $x$, given its encoding and the learnt codebook. Below we discuss two decoding schemes that are diametrically opposite in terms of the inference cost. Also, note that the standard linear classification with real-valued representation and classifiers scale as $O(L)$ in terms of computational complexity and model size.

\subsubsection{Decoding Schemes}

\paragraph{Exact Decoding (ED).} Exact Decoding scheme expects the Hamming distance between the generated instance code, $B(\mathbf{P}F(\mathbf{x};\theta_F))$, and the ground truth class code, $B(\mathbf{C}_i)$ to be {\em exactly} $0$. That is, we can hash the class codes in a table, and then ED requires only a $O(1)$ hash-table lookup for a given instance. Consequently, the inference time for ED is nearly {\em independent} of $L$. Naturally, the decoding scheme is highly stringent and would misclassify an instance if the instance binary code and the ground truth code do not match in even a single {\em bit}. Surprisingly, this highly efficient decoding scheme still provides non-trivial accuracy (see Table~\ref{tab:classification_20bits} and Section~\ref{sec:class_exps}). 

\paragraph{Minimum Hamming Decoding (MHD).} Minimum Hamming Decoding is akin to the Maximum Dot Product used by standard linear classifiers. For an instance code, we evaluate the Hamming distance with all the $L$ class codes and output the class with the least Hamming distance. Note that the Hamming distance over binary codes can be computed using XOR operations that are implemented significantly more efficiently than the floating-point operations~\citep{rastegari2016xnor}. 
Even though, technically, computational complexity and model size of MHD scales as $O(L)$ but the real-world implementations should be an order of magnitude faster than standard classifiers. In fact, for large number of classes $L$, the efficiency of MHD can be further improved by using approximate nearest neighbour search~\citep{datar2004locality,bentley1990k,malkov2018efficient}. Appendix~\ref{sec:alg} has the mathematical presentation of the decoding schemes.

See Section~\ref{sec:disc} for more discussion on potential decoding schemes. Also see Section~\ref{sec:ablations} for ablation studies about the two decoding schemes along with feature separability (linear vs Hamming). 

\subsubsection{Empirical Evaluation}
\label{sec:class_exps}

ImageNet-1K~\citep{russakovsky2015imagenet} is a widely used  image classification dataset with 1000 hierarchical classes. Our classification experiments use ResNet50~\citep{he2016deep} and are trained using the $\sim$1.3M training images. Images were transformed \& augmented with standard procedures~\citep{kusupati2020soft,wortsman2019discovering}. All the implementations were in PyTorch~\citep{paszke2019pytorch} and experimented on a machine with 4 NVIDIA Titan X (Pascal) GPUs. 

When applied to ImageNet-1K, the first phase of \llc,  learnt a 20-bit codebook with 1000 unique class codes, i.e., every class has its own {\em distinct} binary code. We warm start the second phase of \llc by the learnt ResNet50 backbone along with the $20$ dimensional projection layer. See Appendix~\ref{sec:hparams} for the hyperparameter values and other training details.

A key feature of \llc is that it jointly learns both the class codebook as well as instance codes. Several existing techniques decouple this learning process where the codebook is constructed separately and is then used to train the instance codes~\citep{ hsu2009multi,crammer2001algorithmic,allwein2000reducing,escalera2010error,akata2015label,weston2011wsabie}. We evaluate the advantage of the {\em joint} learning approach of \llc by comparing its performance against three strong baselines: i) Random codebook of $20$-bits, ii) $20$-bit CCA codebook~\citep{akata2015label,weston2011wsabie,zhang2011multi} \& iii) $20$-bit SVD codebook. Previous works~\citep{hsu2009multi,crammer2001algorithmic,allwein2000reducing} argued that random codebooks are competitive to the ones constructed using side-information. $20$-bit CCA and SVD codebooks aim to capture the hierarchy that is amiss in the random codebook. The $20$-bit SVD codebook is built using the SVD of $2048$ dimensional linear classifiers (for each class) in the pre-trained ResNet50, and binarizing it. $20$-bit CCA codebook is the binarized version of the transformed label embedding projected on to $20$ components learnt using CCA between 2048 dimensional representations of 50K samples from the ImageNet train set and their one-hot label embeddings. Despite being able to capture the hierarchy information, both $20$-bit CCA/SVD codebooks suffer from clashes reducing their overall effectiveness.

Next, using the baselines codebooks and the corresponding learnt instance codes, we compute class predictions for each test instance using the  Exact Decoding (ED) \& Minimum Hamming Decoding (MHD) schemes mentioned in the previous section. We evaluate all the methods using top-$1$ accuracy on the ImageNet-1K validation set. Baseline ResNet50 architecture represents the maximum accuracy we can hope to achieve using binarized instance$+$class codes. Note that this baseline classifier requires $O(L)$ computation over 16-bit real numbers, and achieves Top-$1$ accuracy of $77\%$. 
\begin{table}[t!]
\centering
\makebox[0pt][c]{\parbox{1\textwidth}{%

    \begin{minipage}[b]{0.52\hsize}
    \centering
    \caption{\small Classification performance on ImageNet-1K  with ResNet50 using various class codebooks for training.}
    \resizebox{0.98\linewidth}{!}
  {
\begin{tabular}{@{}lccc@{}}
\toprule
Codebook                           & \begin{tabular}[c]{@{}c@{}}Unique \\ Codes\end{tabular}  & \begin{tabular}[c]{@{}c@{}}ED \\ Accuracy (\%)\end{tabular}  & \begin{tabular}[c]{@{}c@{}}MHD \\ Accuracy (\%)\end{tabular}  \\ \midrule
Random 20-bits                     & 1000         & 64.07            & 66.91             \\
CCA 20-bits                        & 813          & 55.17            & 57.03             \\
SVD 20-bits                        & 969          & 65.12            & 69.18             \\
\llc 20-bits (Ours) & 1000         & \textbf{68.82}            & \textbf{74.57}             \\ \bottomrule
\end{tabular}

\label{tab:classification_20bits}
  }
    \end{minipage}
    \hfill
    \begin{minipage}[b]{0.46\hsize}
      \centering
    \caption{\small Classification accuracy on ImageNet-1K vs. bit length of the learnt class codebooks (\cref{sec:ablations}).}
    \resizebox{0.98\linewidth}{!}
  {
\begin{tabular}{@{}cccc@{}}
\toprule
\llc Length & \begin{tabular}[c]{@{}c@{}}Unique \\ Codes\end{tabular}  & \begin{tabular}[c]{@{}c@{}}ED \\ Accuracy (\%)\end{tabular}  & \begin{tabular}[c]{@{}c@{}}MHD \\ Accuracy (\%)\end{tabular}  \\ \midrule
15 bits           & 990          & 67.20            & 71.03             \\
20 bits           & 1000         & \textbf{68.82}            & 74.57             \\
25 bits           & 1000         & 67.93            & 74.79             \\
30 bits           & 1000         & 67.51            & \textbf{75.13}             \\ \bottomrule
\end{tabular}

  \label{tab:classification_variousbits}
  }
    \end{minipage}
}}
\end{table}

Table~\ref{tab:classification_20bits} compares the accuracy of \llc (with $20$-bit codebook) against baseline codebooks mentioned above. Note that MHD with \llc codebook is $74.5\%$ accurate, i.e., despite using only $20$-dimensional {\em binary} representation it is only about $2.5\%$ less accurate than standard ResNet50 that uses 2048 dimensional real-valued representation. Furthermore, we observe that compared to standard codebooks like SVD, our jointly learnt codebook is 5\% more accurate. 

Interestingly, Exact Decoding (ED) -- which is $O(1)$ inference scheme -- with \llc codebook is nearly as accurate as the SVD codebook with MHD scheme and is about 12\% more accurate than the CCA codebook with ED scheme. Naturally, codebook length/dimensionality plays a critical role in classification accuracy; see Section~\ref{sec:ablations} for a detailed ablative study on this aspect. Finally, the gains in efficiency should be even more compelling for problems with millions of classes~\citep{varma2019extreme}.

\subsection{Efficient Retrieval}
\label{sec:retrieval_expts}
The goal in retrieval is to find instances from a database that are most similar to a given {\em query}. Traditional retrieval approaches, use a {\em fixed metric} to retrieve "similar points", with data structures like LSH for efficient retrieval. Recent progress in Deep Supervised Hashing (DSH)~\citep{liu2016deep} offer significantly more compelling solutions by learning the hashing function itself. That is, DSH aims to learn binary codes for each instance s.t. a pair of instances are embedded closely iff they belong to the same class, and then learns the hashing function end-to-end using a small train set. 

As \llc also learns instance codes to reflect class membership, we can directly use our learnt encoder as a hashing function for given instances. For each query, the most relevant samples from the database are retrieved based on the minimum Hamming distance. Similar to the decoding schemes in classification, the retrieval can be optimized using approximate nearest neighbor search. Finally, the efficiency gains provided by using bits instead of real numbers should enable deployment of \llc based retrieval for efficient high recall portions of retrieval pipelines. 

\subsubsection{Empirical Evaluation}

Following DSH literature, we evaluate hashing-based image retrieval on ImageNet-100, a benchmark dataset created by~\citet{cao2017hashnet}. ImageNet-100 has 100 classes randomly sampled from ImageNet-1K. All the validation images of these classes are used as query images, all the training images ($\sim$ 1300 per class) of these 100 classes are used as database images. Finally, 130 samples per class from the database are used as the training set for learning binary codes or hashing functions.

We compare against HashNet~\citep{cao2017hashnet} and Greedy Hash~\citep{su2018greedy} for image retrieval using learnt instance codes. HashNet learns the bit representations of instances using a pairwise optimization with positive and negative instance pairs. HashNet is a representative baseline for an alternative way of learning binary instance codes compared to \llc. On the other hand, Greedy Hash learns only the instance codes using straight-through-estimator via the classification task. Note that \llc learns both class codes as well as instance codes differentiating it from Greedy Hash style methods. Learnt instance codes are a byproduct of efficient classification as opposed to baselines that optimize for them.

We use the Mean Average Precision (MAP@1000) metric for evaluation. The MAP@1000 calculation code of HashNet~\citep{cao2017hashnet} is erroneous and has propagated to several papers in the literature. We use the corrected version, hence the accuracy numbers are different from the original paper. Please see Appendix~\ref{sec:metric} for the corrected version, the changes required along with an example and a brief discussion. We used the publicly available pre-trained HashNet models~\citep{hashnetcode} and Greedy Hash~\citep{su2018greedy} code to recompute the MAP@1000. 

Following HashNet~\citep{cao2017hashnet}, we use AlexNet~\citep{krizhevsky2012imagenet} as the backbone and warm-start it with a pre-trained model on ImageNet-1K. We add a projection layer to the backbone and learn the instance and class codes. We also report retrieval numbers with  ResNet50~\citep{he2016deep} and compare \llc based retrieval numbers to learnt real-valued representations. Please see Appendix~\ref{sec:hparams} for the training details and hyperparameters of efficient retrieval pipelines.
\begin{table}[t!]
\centering
\makebox[0pt][c]{\parbox{1\textwidth}{%

    \begin{minipage}[b]{0.51\hsize}
    \centering
    \caption{\small Efficient image retrieval on ImageNet-100 using AlexNet compared using MAP@1000 (Appendix~\ref{sec:metric}) across 16 -- 64 bits. }
    \resizebox{0.98\linewidth}{!}
  {
\begin{tabular}{@{}lccccc@{}}
\toprule
Method              & 10 bits & 16 bits & 32 bits & 48 bits & 64 bits \\ \midrule
HashNet~\citep{cao2017hashnet}             & 0.1995       & 0.2815  & 0.4300  & 0.5270  & 0.5124  \\
Greedy Hash~\citep{su2018greedy}             & 0.2860       & 0.4247  & 0.5412  & 0.5720  & 0.5895  \\
\llc (Ours) & \textbf{0.3086}  & \textbf{0.4305}  & \textbf{0.5565}  & \textbf{0.5749}  & \textbf{0.6000}  \\ \bottomrule
\end{tabular}

\label{tab:hashing_alex}
  }
    \end{minipage}
    \hfill
    \begin{minipage}[b]{0.44\hsize}

      \centering
    \caption{\small  Comparison of \llc based retrieval vs real-valued representations with ResNet50 on ImageNet-100 using MAP@1000.}
    \resizebox{0.98\linewidth}{!}
  {
\begin{tabular}{@{}lccc@{}}
\toprule
Representation & 8 dims & 10 dims & 64 dims \\\midrule
\llc (1 bit)                            & -       & 0.6458        & 0.6773        \\
Real (16 bits)                      & 0.5041        & 0.6657 &        0.7794       \\ \bottomrule
\end{tabular}

  \label{tab:hashing_resnet}
  }
    \end{minipage}
}}
\end{table}

Table~\ref{tab:hashing_alex} shows the performance (evaluated using MAP@1000) for HashNet, Greedy Hash, and \llc across various code lengths. \llc outperforms HashNet across all code lengths (16 -- 64) by at least  4.79\% on MAP@1000. \llc is also better than Greedy Hash across all the bit lengths. \llc also outperforms $16$-bit HashNet by $2\%$ \& $15\%$ using only $10$ \& $16$ bits respectively. Finally, $32$-bit \llc comfortably outperforms both $48$ \& $64$-bit HashNet showcasing the effectiveness of our learnt tight bit codes. Note that \llc, learning both instance and class codes, is effective in retrieval even though it was designed for classification.   

We repeat the retrieval experiments with ResNet50. Table~\ref{tab:hashing_resnet} shows the MAP@1000 for \llc with $10$ and $64$ bits along with the same dimensional real-valued representations. The $10$-bit \llc is only 2\% lower than $10$ dimensional real-valued representation even though theoretically, the cost associated with 10-bit \llc based retrieval is about $256\times$ less than $10$ dimensional real representations. 

The $64$ bit and $10$ bit \llc outperforms $10$ and $8$ dimensional real-valued representations respectively at a much cheaper retrieval cost, at least by an order of magnitude. More discussion about the use of binary codes for retrieval at a large scale can be found in Section~\ref{sec:disc}. Finally, $10$-bit \llc with ResNet50 outperforms the best performing AlexNet based models for the same task, suggesting ResNet50 is a more appropriate architecture for benchmarking DSH literature. 

\subsection{Out-of-Distribution (OOD) Detection}
\label{sec:ood_expts}

For a multi-class classifier, detecting an OOD sample is very important for robustness~\citep{hendrycks2016baseline} and sequential learning~\citep{wallingford2020overfitting}. Multi-class classifiers are augmented with OOD detection capability by setting a threshold on heuristics like maximum logit which is tuned using a validation set. 

We focus on the scenario where the ratio of in-distribution to out-of-distribution samples in the validation set is not representative of the deployment. This throws off the methods that try to maximize metrics, F1, using a validation set. Our learnt class codebook from \llc comes with over-provisioning (for ease of optimization) resulting in unassigned codes. These unassigned codes can be treated as OOD out-of-the-box with no tuning whatsoever. That is, we classify an instance as OOD if its instance code does not match {\em exactly}  with the code of a class in our learnt codebook.

Appendix~\ref{sec:ood_expts_app} discusses the OOD detection experiments on ImageNet-750~\citep{wallingford2020overfitting} \& MIT Places~\citep{zhou2014learning}. At a high level, \llc based out-of-the-box OOD detection (with a learnt 20-bit codebook on ImageNet-1K) achieves nearly the same OOD detection accuracy as a baseline~\citep{hendrycks2016baseline} that tries to maximize F1 using a validation set. We would like to stress that while such a method needs $\approx 3000$ points in the validation set, our method requires {\em no} samples, which is critical in several practical settings.

\subsection{Ablation Studies}
\label{sec:ablations}

\paragraph{Classification Accuracy vs Number of Bits.}

Table~\ref{tab:classification_variousbits} shows the trade-off in classification accuracy with the variation in the length of the learnt codebook for ImageNet-1K. \llc learns a $15$-bit codebook with only 990 unique codes leading to a loss of accuracy due to code collapse in both ED and MHD schemes ($1.62\%$ \& $3.54\%$ compared to 20-bit codebook respectively). An interesting observation is that the ED accuracy gradually goes down after 20-bits while the MHD accuracy keeps on increasing. The phenomenon of increasing accuracy with MHD is probably due to the increase in the capacity of both instance and class codes. However, the decrease in ED accuracy after 20-bits can be explained through the hardness in exactly predicting every bit in the instance code to match the ground truth class code. Our classification model with 20-bits on average gets $19.2$ bits correct but the model with 30-bits only gets $28.5$ bits right. This increase in uncertainty coupled with the stringent ED scheme leads to a slight dip in accuracy as the code length increases. However, this also provides us with a path for more accurate decoding schemes while being efficient as discussed in Section~\ref{sec:disc}.

\paragraph{Classification Accuracy vs Faster Codebook Learning.}

Codebook learning phase of \llc is expensive, this motivated us to speed up codebook learning at a minimal loss in accuracy. One way is to warm-start the codebook using the ones built with SVD/CCA (see Section~\ref{sec:classification_expts}). While these codebooks suffer from code collapse, with further training, they start to learn 1000 unique codes quickly. Using these final codebooks gets to a comparable ($1\%$ drop) accuracy as the $20$-bit learnt \llc codebook but at a relatively cheaper training.  Another option is to use only a portion of the data and a much smaller network to learn the codebook. We sampled 50K training images and use a MobileNetV1~\citep{howard2017mobilenets} (which has about $6\times$ less parameters and compute than ResNet50) to learn a 20-bit codebook which gets to ED and MHD accuracy of $66.62\%$ \& $72.55\%$ which is only $2\%$ lower than the end-to-end learnt codebook but at a fraction of the training cost (3 hrs vs 2 days).

\paragraph{Linear vs Hamming Separability.}
Fitting a deep neural network to the learnt codebook for classification results in warping of the feature space considerably. The final classification space is a hypercube with the vertices being apart by Hamming distance of 1. To verify linear separability, we take the learnt, frozen ResNet50 trained for the 20-bit classification problem and fit a linear classifier on top of the 2048 dimensional features. Linear classifier quickly reaches a top-$1$ accuracy of $75.51\%$.

The opposite does not seem to be true. We extract and freeze the backbone of a pre-trained ResNet50 and train a projection layer to fit the $20$-bit learnt codebook. This gets to top-$1$ accuracy of only about $21\%$ with the ED scheme. However, we also observed that unfreezing and finetuning the last 3 layers of the backbone recovers the top-$1$ ED accuracy to roughly $68\%$.

These experiments show 1) Hamming separability inherently enables linear separability, 2) Linear separability does not imply Hamming separability \& 3) with enough overparameterization, linearly separable space can be warped to support Hamming separability. Hamming separability automatically provides linear separability with increased accuracy of $\sim 1\%$ over the MHD scheme which allows for an option for using a more powerful yet simple classifier, in case of accuracy requirements.

\paragraph{Independent vs Nested Codebook Learning.}

Consider a scenario with varied computational budgets for classification. We could either train independent $k$-bit models (eg., $k=20,25,30$) and use them according to the budget, or we could learn a single nested codebook-based model that can be readily adapted to any of these settings. While training a codebook of larger bit length like $k=30$, we can ensure that the first $m$-bits, $m<k$, also form a codebook at minimal additional cost. We were able to stably train a $30$-bit codebook and also extract $20$, $25$-bit codebooks from it all of which are as accurate as independently trained codebooks. These nested codebooks have the potential to be used based on the computational resource availability for efficient classification without having to retrain.

\section{Discussion and Conclusions}
\label{sec:disc}

We designed \llc to learn low-dimensional binary codes for instances as well as classes, and utilized them in applications like efficient classification, retrieval, OOD detection. A key finding is that combining class code learning with ECOC framework to learn instance code leads to a stable training system that can accurately capture the semantics of the class data despite just $20$-dimensional code.  Traditionally, methods like HashNet, KLSH~\cite{kulis2009kernelized} attempt to learn hashing function using pairwise loss functions by embedding instances such that points from the same class are embedded closely and points from different classes are far. But such formulations are hard to optimize, due to the risk of embedding collapse. We observe that \llc by using instance-wise formulation can train stably with significantly higher performance.   
Another fascinating observation is that while architectures like ResNet50 have large intermediate ($2048$ dimensional real) representations, they can be compressed to just $20$ bits without significant loss in accuracy! Even though quantization~\cite{rastegari2016xnor} literature demonstrates strong compression of representations, we believe such stark compression has been elusive so far and is worth further exploration from the efficient inference viewpoint.

\paragraph{Limitations.}

Our visualization (see Figure~\ref{fig:bit-splits} in Appendix~\ref{sec:quant_hier})  indicates that each bit does not correspond to some easily interpretable attributes, unlike DBC~\citep{rastegari2012attribute}. We believe incorporating priors with weak supervision as well as cross-modal learning could help \llc get past this limitation.

ED and MHD schemes are on two ends of the computation vs accuracy spectrum and do not transition smoothly. Designing decoding schemes that can compromise between these two extreme decoding schemes might be able to address this limitation. 

Strong encoders are needed to warp the feature space to ensure Hamming separability. For example, using a $20$-bit learnt codebook with ED scheme, ResNet50 gets to $68.8\%$ top-$1$ accuracy whereas MobileNetV1 can only reach $53.23\%$. This also ties into the argument that classification is a trade-off between encoder and decoder costs. Making decoders efficient and cheap, puts the burden on encoding the information in the right way and higher expressivity often helps in that cause.

\paragraph{Future Work.}

There are several exciting directions that we would like to explore. In principle, \llc can easily incorporate side-information when needed with simple additional losses during training. The additional regularization losses can also help in incorporating natural constraints on the codebook~\citep{crammer2001algorithmic,allwein2000reducing} or can enable  attribute-based class codes for interpretablity~\citep{farhadi2009describing,ferrari2007learning,lampert2013attribute} making them exciting directions to explore. \llc algorithm can also be used to encode instances of multiple modalities like audio, visual, language to the same learnt low-dimensional binary space. This might help in effective cross-modal supervision along with retrieval among various other applications.

\paragraph{Potential in Large-Scale Applications.}

While our focus was on designing low-dimensional accurate binary codes, several studies~\citep{rastegari2016xnor,hubara2016binarized} have shown that efficiency afforded by bit-wise computation over floating-point computation can lead to almost an order of magnitude speed-up. Furthermore, as the number of classes increases, the learning of class codebooks helps in training representations in sublinear costs~\citep{cisse2013robust} along with sublinear inference (in $L$). We expect \llc algorithm to have its efficiency benefits outweigh the accuracy drop for large multi-class/multi-label problems, like objection recognition using ImageNet-22K~\citep{deng2009imagenet}, document tagging~\citep{varma2019extreme,prabhu2020extreme} and instance classification~\citep{weyand2020google}. The efficiency aspect of the binary codes has not been fully explored in this paper as the main computational bottleneck for ImageNet-1K classification is the deep neural network featurizer. 

Lastly, \llc based efficient retrieval can be used for the initial high-recall shortlisting of a search pipeline, which is followed by high precision models operating on more expressive yet expensive embeddings. We leave practical demonstration of such a system at web-scale for future work.

\section*{Acknowledgments}
We are grateful to Tapan Chugh, Kunal Dahiya, Max Horton, Sewoong Oh, Mohammad Rastegari, Ludwig Schmidt and members of RAIVN Lab for helpful discussions and feedback. AK would also like to thank Soumen Chakrabarti and Manik Varma for sowing the seeds of this idea in his initial research days. Sham Kakade acknowledges funding from NSF Awards CCF-1703574. Ali Farhadi acknowledges funding from the NSF awards IIS 1652052, IIS 1703166, DARPA N66001-19-2-4031, W911NF-15-1-0543 and gifts from Allen Institute for Artificial Intelligence.
\bibliography{local}

\begin{thebibliography}{65}
\providecommand{\natexlab}[1]{#1}
\providecommand{\url}[1]{\texttt{#1}}
\expandafter\ifx\csname urlstyle\endcsname\relax
  \providecommand{\doi}[1]{doi: #1}\else
  \providecommand{\doi}{doi: \begingroup \urlstyle{rm}\Url}\fi

\bibitem[Akata et~al.(2015{\natexlab{a}})Akata, Perronnin, Harchaoui, and
  Schmid]{akata2015label}
Z.~Akata, F.~Perronnin, Z.~Harchaoui, and C.~Schmid.
\newblock Label-embedding for image classification.
\newblock \emph{IEEE transactions on pattern analysis and machine
  intelligence}, 38\penalty0 (7):\penalty0 1425--1438, 2015{\natexlab{a}}.

\bibitem[Akata et~al.(2015{\natexlab{b}})Akata, Reed, Walter, Lee, and
  Schiele]{akata2015evaluation}
Z.~Akata, S.~Reed, D.~Walter, H.~Lee, and B.~Schiele.
\newblock Evaluation of output embeddings for fine-grained image
  classification.
\newblock In \emph{Proceedings of the IEEE conference on computer vision and
  pattern recognition}, pages 2927--2936, 2015{\natexlab{b}}.

\bibitem[Allwein et~al.(2000)Allwein, Schapire, and
  Singer]{allwein2000reducing}
E.~L. Allwein, R.~E. Schapire, and Y.~Singer.
\newblock Reducing multiclass to binary: A unifying approach for margin
  classifiers.
\newblock \emph{Journal of machine learning research}, 1\penalty0
  (Dec):\penalty0 113--141, 2000.

\bibitem[Bautista~Mart{\'\i}n et~al.(2016)]{bautista2016learning}
M.~{\'A}. Bautista~Mart{\'\i}n et~al.
\newblock \emph{Learning error-correcting representations for multi-class
  problems}.
\newblock PhD thesis, Universitat de Barcelona, 2016.

\bibitem[Bengio et~al.(2010)Bengio, Weston, and Grangier]{bengio2010label}
S.~Bengio, J.~Weston, and D.~Grangier.
\newblock Label embedding trees for large multi-class tasks.
\newblock In \emph{Proceedings of the 23rd International Conference on Neural
  Information Processing Systems-Volume 1}, pages 163--171, 2010.

\bibitem[Bengio et~al.(2013)Bengio, L{\'e}onard, and
  Courville]{bengio2013estimating}
Y.~Bengio, N.~L{\'e}onard, and A.~Courville.
\newblock Estimating or propagating gradients through stochastic neurons for
  conditional computation.
\newblock \emph{arXiv preprint arXiv:1308.3432}, 2013.

\bibitem[Bentley(1990)]{bentley1990k}
J.~L. Bentley.
\newblock K-d trees for semidynamic point sets.
\newblock In \emph{Proceedings of the sixth annual symposium on Computational
  geometry}, pages 187--197, 1990.

\bibitem[Bhatia et~al.(2015)Bhatia, Jain, Kar, Varma, and
  Jain]{bhatia2015sparse}
K.~Bhatia, H.~Jain, P.~Kar, M.~Varma, and P.~Jain.
\newblock Sparse local embeddings for extreme multi-label classification.
\newblock In \emph{NIPS}, volume~29, pages 730--738, 2015.

\bibitem[Cao et~al.(2017{\natexlab{a}})Cao, Long, Wang, and Yu]{cao2017hashnet}
Z.~Cao, M.~Long, J.~Wang, and P.~S. Yu.
\newblock Hashnet: Deep learning to hash by continuation.
\newblock In \emph{Proceedings of the IEEE international conference on computer
  vision}, pages 5608--5617, 2017{\natexlab{a}}.

\bibitem[Cao et~al.(2017{\natexlab{b}})Cao, Long, Wang, and Yu]{hashnetcode}
Z.~Cao, M.~Long, J.~Wang, and P.~S. Yu.
\newblock {Hashnet: Deep learning to hash by continuation}, 2017{\natexlab{b}}.
\newblock URL \url{https://github.com/thuml/HashNet}.

\bibitem[Carreira-Perpin{\'a}n and Raziperchikolaei(2015)]{carreira2015hashing}
M.~A. Carreira-Perpin{\'a}n and R.~Raziperchikolaei.
\newblock Hashing with binary autoencoders.
\newblock In \emph{Proceedings of the IEEE conference on computer vision and
  pattern recognition}, pages 557--566, 2015.

\bibitem[Choromanska et~al.(2016)Choromanska, Choromanski, Bojarski, Jebara,
  Kumar, and LeCun]{choromanska2016binary}
A.~Choromanska, K.~Choromanski, M.~Bojarski, T.~Jebara, S.~Kumar, and Y.~LeCun.
\newblock Binary embeddings with structured hashed projections.
\newblock In \emph{International Conference on Machine Learning}, pages
  344--353. PMLR, 2016.

\bibitem[Ciss{\'e} et~al.(2013)Ciss{\'e}, Usunier, Artieres, and
  Gallinari]{cisse2013robust}
M.~Ciss{\'e}, N.~Usunier, T.~Artieres, and P.~Gallinari.
\newblock Robust bloom filters for large multilabel classification tasks.
\newblock In \emph{Advances in Neural Information Processing Systems 26}, pages
  1851--1859, 2013.

\bibitem[Crammer and Singer(2001)]{crammer2001algorithmic}
K.~Crammer and Y.~Singer.
\newblock On the algorithmic implementation of multiclass kernel-based vector
  machines.
\newblock \emph{Journal of machine learning research}, 2\penalty0
  (Dec):\penalty0 265--292, 2001.

\bibitem[Dahiya et~al.(2021)Dahiya, Saini, Mittal, Shaw, Dave, Soni, Jain,
  Agarwal, and Varma]{dahiya2021deepxml}
K.~Dahiya, D.~Saini, A.~Mittal, A.~Shaw, K.~Dave, A.~Soni, H.~Jain, S.~Agarwal,
  and M.~Varma.
\newblock Deepxml: A deep extreme multi-label learning framework applied to
  short text documents.
\newblock In \emph{Proceedings of the 14th ACM International Conference on Web
  Search and Data Mining}, pages 31--39, 2021.

\bibitem[Datar et~al.(2004)Datar, Immorlica, Indyk, and
  Mirrokni]{datar2004locality}
M.~Datar, N.~Immorlica, P.~Indyk, and V.~S. Mirrokni.
\newblock Locality-sensitive hashing scheme based on p-stable distributions.
\newblock In \emph{Proceedings of the twentieth annual symposium on
  Computational geometry}, pages 253--262, 2004.

\bibitem[Deng et~al.(2009)Deng, Dong, Socher, Li, Li, and
  Fei-Fei]{deng2009imagenet}
J.~Deng, W.~Dong, R.~Socher, L.-J. Li, K.~Li, and L.~Fei-Fei.
\newblock Imagenet: A large-scale hierarchical image database.
\newblock In \emph{2009 IEEE conference on computer vision and pattern
  recognition}, pages 248--255. Ieee, 2009.

\bibitem[Deng et~al.(2011)Deng, Berg, and Fei-Fei]{deng2011hierarchical}
J.~Deng, A.~C. Berg, and L.~Fei-Fei.
\newblock Hierarchical semantic indexing for large scale image retrieval.
\newblock In \emph{CVPR 2011}, pages 785--792. IEEE, 2011.

\bibitem[Dietterich and Bakiri(1994)]{dietterich1994solving}
T.~G. Dietterich and G.~Bakiri.
\newblock Solving multiclass learning problems via error-correcting output
  codes.
\newblock \emph{Journal of artificial intelligence research}, 2:\penalty0
  263--286, 1994.

\bibitem[Escalera et~al.(2010)Escalera, Pujol, and Radeva]{escalera2010error}
S.~Escalera, O.~Pujol, and P.~Radeva.
\newblock Error-correcting ouput codes library.
\newblock \emph{The Journal of Machine Learning Research}, 11:\penalty0
  661--664, 2010.

\bibitem[Farhadi et~al.(2009)Farhadi, Endres, Hoiem, and
  Forsyth]{farhadi2009describing}
A.~Farhadi, I.~Endres, D.~Hoiem, and D.~Forsyth.
\newblock Describing objects by their attributes.
\newblock In \emph{2009 IEEE Conference on Computer Vision and Pattern
  Recognition}, pages 1778--1785. IEEE, 2009.

\bibitem[Ferrari and Zisserman(2007)]{ferrari2007learning}
V.~Ferrari and A.~Zisserman.
\newblock Learning visual attributes.
\newblock \emph{Advances in neural information processing systems},
  20:\penalty0 433--440, 2007.

\bibitem[Gong et~al.(2012)Gong, Lazebnik, Gordo, and
  Perronnin]{gong2012iterative}
Y.~Gong, S.~Lazebnik, A.~Gordo, and F.~Perronnin.
\newblock Iterative quantization: A procrustean approach to learning binary
  codes for large-scale image retrieval.
\newblock \emph{IEEE transactions on pattern analysis and machine
  intelligence}, 35\penalty0 (12):\penalty0 2916--2929, 2012.

\bibitem[Guo et~al.(2019)Guo, Mousavi, Wu, Holtmann-Rice, Kale, Reddi, and
  Kumar]{Guo2019BreakingTG}
C.~Guo, A.~Mousavi, X.~Wu, D.~Holtmann-Rice, S.~Kale, S.~J. Reddi, and
  S.~Kumar.
\newblock Breaking the glass ceiling for embedding-based classifiers for large
  output spaces.
\newblock In \emph{NeurIPS}, 2019.

\bibitem[He et~al.(2016)He, Zhang, Ren, and Sun]{he2016deep}
K.~He, X.~Zhang, S.~Ren, and J.~Sun.
\newblock Deep residual learning for image recognition.
\newblock In \emph{Proceedings of the IEEE conference on computer vision and
  pattern recognition}, pages 770--778, 2016.

\bibitem[Hendrycks and Gimpel(2016)]{hendrycks2016baseline}
D.~Hendrycks and K.~Gimpel.
\newblock A baseline for detecting misclassified and out-of-distribution
  examples in neural networks.
\newblock \emph{arXiv preprint arXiv:1610.02136}, 2016.

\bibitem[Howard et~al.(2017)Howard, Zhu, Chen, Kalenichenko, Wang, Weyand,
  Andreetto, and Adam]{howard2017mobilenets}
A.~G. Howard, M.~Zhu, B.~Chen, D.~Kalenichenko, W.~Wang, T.~Weyand,
  M.~Andreetto, and H.~Adam.
\newblock Mobilenets: Efficient convolutional neural networks for mobile vision
  applications.
\newblock \emph{arXiv preprint arXiv:1704.04861}, 2017.

\bibitem[Hsu et~al.(2009)Hsu, Kakade, Langford, and Zhang]{hsu2009multi}
D.~Hsu, S.~M. Kakade, J.~Langford, and T.~Zhang.
\newblock Multi-label prediction via compressed sensing.
\newblock \emph{arXiv preprint arXiv:0902.1284}, 2009.

\bibitem[Hubara et~al.(2016)Hubara, Courbariaux, Soudry, El-Yaniv, and
  Bengio]{hubara2016binarized}
I.~Hubara, M.~Courbariaux, D.~Soudry, R.~El-Yaniv, and Y.~Bengio.
\newblock Binarized neural networks.
\newblock In \emph{Proceedings of the 30th International Conference on Neural
  Information Processing Systems}, pages 4114--4122, 2016.

\bibitem[Jain et~al.(2019)Jain, Balasubramanian, Chunduri, and
  Varma]{jain2019slice}
H.~Jain, V.~Balasubramanian, B.~Chunduri, and M.~Varma.
\newblock Slice: Scalable linear extreme classifiers trained on 100 million
  labels for related searches.
\newblock In \emph{Proceedings of the Twelfth ACM International Conference on
  Web Search and Data Mining}, pages 528--536, 2019.

\bibitem[Krizhevsky et~al.(2012)Krizhevsky, Sutskever, and
  Hinton]{krizhevsky2012imagenet}
A.~Krizhevsky, I.~Sutskever, and G.~E. Hinton.
\newblock Imagenet classification with deep convolutional neural networks.
\newblock \emph{Advances in neural information processing systems},
  25:\penalty0 1097--1105, 2012.

\bibitem[Kulis and Darrell(2009)]{kulis2009learning}
B.~Kulis and T.~Darrell.
\newblock Learning to hash with binary reconstructive embeddings.
\newblock In \emph{NIPS}, volume~22, pages 1042--1050. Citeseer, 2009.

\bibitem[Kulis and Grauman(2009)]{kulis2009kernelized}
B.~Kulis and K.~Grauman.
\newblock Kernelized locality-sensitive hashing for scalable image search.
\newblock In \emph{2009 IEEE 12th international conference on computer vision},
  pages 2130--2137. IEEE, 2009.

\bibitem[Kulis et~al.(2009)Kulis, Jain, and Grauman]{kulis2009fast}
B.~Kulis, P.~Jain, and K.~Grauman.
\newblock Fast similarity search for learned metrics.
\newblock \emph{IEEE Transactions on Pattern Analysis and Machine
  Intelligence}, 31\penalty0 (12):\penalty0 2143--2157, 2009.

\bibitem[Kusupati et~al.(2020)Kusupati, Ramanujan, Somani, Wortsman, Jain,
  Kakade, and Farhadi]{kusupati2020soft}
A.~Kusupati, V.~Ramanujan, R.~Somani, M.~Wortsman, P.~Jain, S.~Kakade, and
  A.~Farhadi.
\newblock Soft threshold weight reparameterization for learnable sparsity.
\newblock In \emph{International Conference on Machine Learning}, pages
  5544--5555. PMLR, 2020.

\bibitem[Lampert et~al.(2009)Lampert, Nickisch, and
  Harmeling]{lampert2009learning}
C.~H. Lampert, H.~Nickisch, and S.~Harmeling.
\newblock Learning to detect unseen object classes by between-class attribute
  transfer.
\newblock In \emph{2009 IEEE Conference on Computer Vision and Pattern
  Recognition}, pages 951--958. IEEE, 2009.

\bibitem[Lampert et~al.(2013)Lampert, Nickisch, and
  Harmeling]{lampert2013attribute}
C.~H. Lampert, H.~Nickisch, and S.~Harmeling.
\newblock Attribute-based classification for zero-shot visual object
  categorization.
\newblock \emph{IEEE transactions on pattern analysis and machine
  intelligence}, 36\penalty0 (3):\penalty0 453--465, 2013.

\bibitem[Liu et~al.(2016)Liu, Wang, Shan, and Chen]{liu2016deep}
H.~Liu, R.~Wang, S.~Shan, and X.~Chen.
\newblock Deep supervised hashing for fast image retrieval.
\newblock In \emph{Proceedings of the IEEE conference on computer vision and
  pattern recognition}, pages 2064--2072, 2016.

\bibitem[Luo et~al.(2020)Luo, Chen, Zhong, Zhang, Deng, Huang, and
  Hua]{luo2020survey}
X.~Luo, C.~Chen, H.~Zhong, H.~Zhang, M.~Deng, J.~Huang, and X.~Hua.
\newblock A survey on deep hashing methods.
\newblock \emph{arXiv preprint arXiv:2003.03369}, 2020.

\bibitem[Malkov and Yashunin(2018)]{malkov2018efficient}
Y.~A. Malkov and D.~A. Yashunin.
\newblock Efficient and robust approximate nearest neighbor search using
  hierarchical navigable small world graphs.
\newblock \emph{IEEE transactions on pattern analysis and machine
  intelligence}, 42\penalty0 (4):\penalty0 824--836, 2018.

\bibitem[Miller(1995)]{miller1995wordnet}
G.~A. Miller.
\newblock Wordnet: a lexical database for english.
\newblock \emph{Communications of the ACM}, 38\penalty0 (11):\penalty0 39--41,
  1995.

\bibitem[Murtagh and Contreras(2012)]{murtagh2012algorithms}
F.~Murtagh and P.~Contreras.
\newblock Algorithms for hierarchical clustering: an overview.
\newblock \emph{Wiley Interdisciplinary Reviews: Data Mining and Knowledge
  Discovery}, 2\penalty0 (1):\penalty0 86--97, 2012.

\bibitem[Norouzi et~al.(2012)Norouzi, Fleet, and
  Salakhutdinov]{norouzi2012hamming}
M.~Norouzi, D.~J. Fleet, and R.~R. Salakhutdinov.
\newblock Hamming distance metric learning.
\newblock In \emph{Advances in neural information processing systems}, pages
  1061--1069, 2012.

\bibitem[Norouzi et~al.(2013)Norouzi, Mikolov, Bengio, Singer, Shlens, Frome,
  Corrado, and Dean]{norouzi2013zero}
M.~Norouzi, T.~Mikolov, S.~Bengio, Y.~Singer, J.~Shlens, A.~Frome, G.~S.
  Corrado, and J.~Dean.
\newblock Zero-shot learning by convex combination of semantic embeddings.
\newblock \emph{arXiv preprint arXiv:1312.5650}, 2013.

\bibitem[Paszke et~al.(2019)Paszke, Gross, Massa, Lerer, Bradbury, Chanan,
  Killeen, Lin, Gimelshein, Antiga, et~al.]{paszke2019pytorch}
A.~Paszke, S.~Gross, F.~Massa, A.~Lerer, J.~Bradbury, G.~Chanan, T.~Killeen,
  Z.~Lin, N.~Gimelshein, L.~Antiga, et~al.
\newblock Pytorch: An imperative style, high-performance deep learning library.
\newblock In \emph{Advances in Neural Information Processing Systems}, pages
  8024--8035, 2019.

\bibitem[Prabhu et~al.(2020)Prabhu, Kusupati, Gupta, and
  Varma]{prabhu2020extreme}
Y.~Prabhu, A.~Kusupati, N.~Gupta, and M.~Varma.
\newblock Extreme regression for dynamic search advertising.
\newblock In \emph{Proceedings of the 13th International Conference on Web
  Search and Data Mining}, pages 456--464, 2020.

\bibitem[Rastegari et~al.(2012)Rastegari, Farhadi, and
  Forsyth]{rastegari2012attribute}
M.~Rastegari, A.~Farhadi, and D.~Forsyth.
\newblock Attribute discovery via predictable discriminative binary codes.
\newblock In \emph{European Conference on Computer Vision}, pages 876--889.
  Springer, 2012.

\bibitem[Rastegari et~al.(2016)Rastegari, Ordonez, Redmon, and
  Farhadi]{rastegari2016xnor}
M.~Rastegari, V.~Ordonez, J.~Redmon, and A.~Farhadi.
\newblock Xnor-net: Imagenet classification using binary convolutional neural
  networks.
\newblock In \emph{European conference on computer vision}, pages 525--542.
  Springer, 2016.

\bibitem[Russakovsky et~al.(2015)Russakovsky, Deng, Su, Krause, Satheesh, Ma,
  Huang, Karpathy, Khosla, Bernstein, et~al.]{russakovsky2015imagenet}
O.~Russakovsky, J.~Deng, H.~Su, J.~Krause, S.~Satheesh, S.~Ma, Z.~Huang,
  A.~Karpathy, A.~Khosla, M.~Bernstein, et~al.
\newblock Imagenet large scale visual recognition challenge.
\newblock \emph{International journal of computer vision}, 115\penalty0
  (3):\penalty0 211--252, 2015.

\bibitem[Salakhutdinov and Hinton(2007)]{salakhutdinov2007learning}
R.~Salakhutdinov and G.~Hinton.
\newblock Learning a nonlinear embedding by preserving class neighbourhood
  structure.
\newblock In \emph{Artificial Intelligence and Statistics}, pages 412--419.
  PMLR, 2007.

\bibitem[Salakhutdinov and Hinton(2009)]{salakhutdinov2009semantic}
R.~Salakhutdinov and G.~Hinton.
\newblock Semantic hashing.
\newblock \emph{International Journal of Approximate Reasoning}, 50\penalty0
  (7):\penalty0 969--978, 2009.

\bibitem[Shen et~al.(2018)Shen, Xu, Liu, Yang, Huang, and
  Shen]{shen2018unsupervised}
F.~Shen, Y.~Xu, L.~Liu, Y.~Yang, Z.~Huang, and H.~T. Shen.
\newblock Unsupervised deep hashing with similarity-adaptive and discrete
  optimization.
\newblock \emph{IEEE transactions on pattern analysis and machine
  intelligence}, 40\penalty0 (12):\penalty0 3034--3044, 2018.

\bibitem[Su et~al.(2018)Su, Zhang, Han, and Tian]{su2018greedy}
S.~Su, C.~Zhang, K.~Han, and Y.~Tian.
\newblock Greedy hash: Towards fast optimization for accurate hash coding in
  cnn.
\newblock In \emph{Proceedings of the 32nd International Conference on Neural
  Information Processing Systems}, pages 806--815, 2018.

\bibitem[Varma(2019)]{varma2019extreme}
M.~Varma.
\newblock Extreme classification.
\newblock \emph{Communications of the ACM}, 62\penalty0 (11):\penalty0 44--45,
  2019.

\bibitem[Wallingford et~al.(2020)Wallingford, Kusupati, Alizadeh-Vahid,
  Walsman, Kembhavi, and Farhadi]{wallingford2020overfitting}
M.~Wallingford, A.~Kusupati, K.~Alizadeh-Vahid, A.~Walsman, A.~Kembhavi, and
  A.~Farhadi.
\newblock Are we overfitting to experimental setups in recognition?
\newblock \emph{arXiv preprint arXiv:2007.02519}, 2020.

\bibitem[Wang et~al.(2015)Wang, Liu, Kumar, and Chang]{wang2015learning}
J.~Wang, W.~Liu, S.~Kumar, and S.-F. Chang.
\newblock Learning to hash for indexing big data—a survey.
\newblock \emph{Proceedings of the IEEE}, 104\penalty0 (1):\penalty0 34--57,
  2015.

\bibitem[Weiss et~al.(2008)Weiss, Torralba, Fergus, et~al.]{weiss2008spectral}
Y.~Weiss, A.~Torralba, R.~Fergus, et~al.
\newblock Spectral hashing.
\newblock In \emph{Advances in Neural Information Processing Systems}, 2008.

\bibitem[Weston et~al.(2011)Weston, Bengio, and Usunier]{weston2011wsabie}
J.~Weston, S.~Bengio, and N.~Usunier.
\newblock Wsabie: Scaling up to large vocabulary image annotation.
\newblock In \emph{Twenty-Second International Joint Conference on Artificial
  Intelligence}, 2011.

\bibitem[Weyand et~al.(2020)Weyand, Araujo, Cao, and Sim]{weyand2020google}
T.~Weyand, A.~Araujo, B.~Cao, and J.~Sim.
\newblock Google landmarks dataset v2-a large-scale benchmark for
  instance-level recognition and retrieval.
\newblock In \emph{Proceedings of the IEEE/CVF Conference on Computer Vision
  and Pattern Recognition}, pages 2575--2584, 2020.

\bibitem[Wortsman et~al.(2019)Wortsman, Farhadi, and
  Rastegari]{wortsman2019discovering}
M.~Wortsman, A.~Farhadi, and M.~Rastegari.
\newblock Discovering neural wirings.
\newblock \emph{arXiv preprint arXiv:1906.00586}, 2019.

\bibitem[Yu et~al.(2014)Yu, Jain, Kar, and Dhillon]{yu2014large}
H.-F. Yu, P.~Jain, P.~Kar, and I.~Dhillon.
\newblock Large-scale multi-label learning with missing labels.
\newblock In \emph{International conference on machine learning}, pages
  593--601. PMLR, 2014.

\bibitem[Yuan et~al.(2018)Yuan, Ren, Lu, and Zhou]{yuan2018relaxation}
X.~Yuan, L.~Ren, J.~Lu, and J.~Zhou.
\newblock Relaxation-free deep hashing via policy gradient.
\newblock In \emph{Proceedings of the European Conference on Computer Vision
  (ECCV)}, pages 134--150, 2018.

\bibitem[Zhang(2020)]{zhang2020survey}
X.~Zhang.
\newblock A survey on deep hashing for image retrieval.
\newblock \emph{arXiv preprint arXiv:2006.05627}, 2020.

\bibitem[Zhang and Schneider(2011)]{zhang2011multi}
Y.~Zhang and J.~Schneider.
\newblock Multi-label output codes using canonical correlation analysis.
\newblock In \emph{Proceedings of the fourteenth international conference on
  artificial intelligence and statistics}, pages 873--882. JMLR Workshop and
  Conference Proceedings, 2011.

\bibitem[Zhou et~al.(2014)Zhou, Lapedriza, Xiao, Torralba, and
  Oliva]{zhou2014learning}
B.~Zhou, A.~Lapedriza, J.~Xiao, A.~Torralba, and A.~Oliva.
\newblock Learning deep features for scene recognition using places database.
\newblock In \emph{Advances in Neural Information Processing Systems}, 2014.

\end{thebibliography}
\section*{Checklist}


\begin{enumerate}

\item For all authors...
\begin{enumerate}
  \item Do the main claims made in the abstract and introduction accurately reflect the paper's contributions and scope?
    \answerYes{}
  \item Did you describe the limitations of your work?
    \answerYes{See Section~\ref{sec:disc}.}
  \item Did you discuss any potential negative societal impacts of your work?
    \answerNA{As with all the efficient machine learning work, this work has both pros and cons. We believe that making ML models affordably available will help in a better feedback loop overall.}
  \item Have you read the ethics review guidelines and ensured that your paper conforms to them?
    \answerYes{}
\end{enumerate}

\item If you are including theoretical results...
\begin{enumerate}
  \item Did you state the full set of assumptions of all theoretical results?
    \answerNA{}
	\item Did you include complete proofs of all theoretical results?
    \answerNA{}
\end{enumerate}

\item If you ran experiments...
\begin{enumerate}
  \item Did you include the code, data, and instructions needed to reproduce the main experimental results (either in the supplemental material or as a URL)?
    \answerYes{See supplemental material. All the code and models will be open sourced.}
  \item Did you specify all the training details (e.g., data splits, hyperparameters, how they were chosen)?
    \answerYes{See Section~\ref{sec:apps+exps} and Appendix~\ref{sec:hparams}.}
	\item Did you report error bars (e.g., with respect to the random seed after running experiments multiple times)?
    \answerNo{We evaluated our method on ImageNet which is computationally expensive to run (3 days per experiment).}
	\item Did you include the total amount of compute and the type of resources used (e.g., type of GPUs, internal cluster, or cloud provider)?
    \answerYes{See Section~\ref{sec:apps+exps}.}
\end{enumerate}

\item If you are using existing assets (e.g., code, data, models) or curating/releasing new assets...
\begin{enumerate}
  \item If your work uses existing assets, did you cite the creators?
    \answerYes{}
  \item Did you mention the license of the assets?
    \answerNo{All the datasets and code used are public and are under MIT, BSD or CC licenses.}
  \item Did you include any new assets either in the supplemental material or as a URL?
    \answerNo{}
  \item Did you discuss whether and how consent was obtained from people whose data you're using/curating?
    \answerNA{}{}
  \item Did you discuss whether the data you are using/curating contains personally identifiable information or offensive content?
    \answerNA{}
\end{enumerate}

\item If you used crowdsourcing or conducted research with human subjects...
\begin{enumerate}
  \item Did you include the full text of instructions given to participants and screenshots, if applicable?
    \answerNA{}
  \item Did you describe any potential participant risks, with links to Institutional Review Board (IRB) approvals, if applicable?
    \answerNA{}
  \item Did you include the estimated hourly wage paid to participants and the total amount spent on participant compensation?
    \answerNA{}
\end{enumerate}

\end{enumerate}


\clearpage
\appendix
\section{\llc Decoding Schemes for Classification \& Binarization Function}
\label{sec:alg}
\begin{algorithm}[ht]
\caption{Inference using Exact Decoding (ED)}
\label{alg:ed}
\begin{algorithmic}[1]
\Require $x\in{\cal X}$, $F(\, \cdot \,; \theta_F)$, $\mathbf{C}$ and $\mathbf{P}$
\Ensure $\ell^*\subseteq [L]$
\State $g(x) \gets B(\mathbf{P}\cdot F(x;\theta_F))$
\State $\ell^* \gets \left\lbrace \ell\in[L] \colon B(\mathbf{C}_\ell) = g(x) \right\rbrace$
\end{algorithmic}
\end{algorithm}

\begin{algorithm}[ht]
\caption{Inference using Minimum Hamming Decoding (MHD)}
\label{alg:mhd}
\begin{algorithmic}[1]
\Require $x\in{\cal X}$, $F(\, \cdot \,; \theta_F)$, $\mathbf{C}$ and $\mathbf{P}$
\Ensure $\ell^* \subseteq [L]$\ \ ($\ell^*\neq \emptyset$)
\State $g(x) \gets B(\mathbf{P}\cdot F(x;\theta_F))$
\State $\ell^* \gets \mathop{\arg\min}_{\ell\in[L]} \frac{1}{2}\lVert B(\mathbf{C}_\ell) - g(x) \rVert_1$
\end{algorithmic}
\end{algorithm}

\begin{algorithm}[!ht]
\caption{PyTorch code for Binarization $B(\cdot)$ function with straight-through-estimator (STE).}
\label{alg:code}

\definecolor{codeblue}{rgb}{0.25,0.5,0.5}
\definecolor{codeblue2}{rgb}{0,0,1}
\lstset{
  backgroundcolor=\color{white},
  basicstyle=\fontsize{9pt}{9pt}\ttfamily\selectfont,
  columns=fullflexible,
  breaklines=true,
  captionpos=b,
  commentstyle=\fontsize{7.2pt}{7.2pt}\color{codeblue},
  keywordstyle=\fontsize{7.2pt}{7.2pt}\color{codeblue2},
}
\begin{lstlisting}[language=python]
class Binarize(autograd.Function):
    @staticmethod
    def forward(ctx, weight):
        out = weight.clone()
        # binarizing in the forward pass
        out[out >= 0] = 1
        out[out < 0] = -1
        
        return out

    @staticmethod
    def backward(ctx, g):
        # send the gradient g straight-through on the backward pass.
        return g, None
        
\end{lstlisting}
\end{algorithm}

\section{Corrected MAP@1000 Metric for Image Retrieval.}
\label{sec:metric}

The reported MAP@1000 metric in~\citet{cao2017hashnet} has an error that results in the wrong estimation of retrieval performance of the learnt hash functions. This error has propagated into many of the follow-up papers rendering the MAP numbers presented in them non-transferable. Most of the papers after HashNet have continued to use the same metric resulting in this situation. The code for the metric is provided as part of the open-sourced codebase of HashNet\footnote{\url{https://github.com/thuml/HashNet/blob/master/caffe/models/predict/imagenet/predict_parallel.py\#L20}}.

Everything until the computation of Precision@k ($k \in [1000]$) is correct. However, the reported metric computes AP@1000 as follows:

$$\text{AP}@1000 = \frac{\sum_k P@k* rel(k)}{\sum_k rel(k)}$$ where $rel(k)$ is an indicator function if the sample at $k$-th position is relevant. ${\sum_k rel(k)}$ is the total number of relevant samples in all the 1000 retrieved samples. It should be noted that every query has around 1300 relevant documents and all of them can not be retrieved within 1000. MAP@k is just the mean of all the AP@k for all the queries. 
\begin{table}[t!]
  \centering
  \caption{\small Image retrieval performance on ImageNet-100 using AlexNet. RMAP: Reported MAP@1000 as in HashNet~\citep{cao2017hashnet}. CMAP: Corrected MAP@1000. Please see Appendix~\ref{sec:metric} for the details. RMAP and CMAP are not well correlated making the reported numbers from literature hard to compare.}
\resizebox{\columnwidth}{!}{%
\begin{tabular}{@{}lcc|cc|cc|cc|cc@{}}
\toprule
\multirow{2}{*}{Method} & \multicolumn{2}{c|}{10 bits} & \multicolumn{2}{c|}{16 bits} & \multicolumn{2}{c|}{32 bits} & \multicolumn{2}{c|}{48 bits} & \multicolumn{2}{c}{64 bits} \\ \cmidrule(l){2-11} 
                        & RMAP          & CMAP         & RMAP          & CMAP         & RMAP          & CMAP         & RMAP          & CMAP         & RMAP         & CMAP         \\ \midrule
HashNet~\citep{cao2017hashnet}                 & 0.3721             & 0.1995            & 0.4634        & 0.2815       & 0.5915        & 0.4300         & 0.6548        & 0.5270        & 0.6542       & 0.5124       \\
\llc (Ours)                     & \textbf{0.4815}        & \textbf{0.3086}       & \textbf{0.5617}        & \textbf{0.4305}       & \textbf{0.6587}        & \textbf{0.5565}       & \textbf{0.6750}        & \textbf{0.5749}       & \textbf{0.6932}       & \textbf{0.6000}       \\ \bottomrule
\end{tabular}
}
\label{tab:hashing_error}
\end{table}
We explain the error using a simple example. Suppose a retrieval problem allows to retrieve 5 samples from a database where each query has 10 relevant samples. And for a given query, let us say method 1 gives retrieves the top-5 samples with the following relevance $\mathbf{[1, 0, 0, 0, 0]}$. And method 2 retrieved with samples with the relevance $\mathbf{[1, 0, 0, 1, 1]}$.

Using the reported metric, method 1 has an AP@5 of $1.0$ (with just 1 relevant sample out of 5). But method 2 with 2 more relevant samples along with the top retrieved sample and has an AP@5 of 0.7. Even with an objectively better retrieval, method 2 has a lower AP@5 score than method 1. This results in an unfair evaluation of methods where methods with poor recall but a few precise retrievals will outperform methods that have both high precision and recall. 

The fix is simple and is just the use of standard MAP@k metric~\footnote{\url{http://sdsawtelle.github.io/blog/output/mean-average-precision-MAP-for-recommender-systems.html\#Average-Precision}}\footnote{\url{https://en.wikipedia.org/wiki/Evaluation_measures_(information_retrieval)\#Average_precision}}. 

$$\text{AP}@1000 = \frac{\sum_k P@k* rel(k)}{\min(\text{total relevant samples}, 1000)}.$$

The reason we need the ${\min(\text{total relevant samples}, 1000)}$ term in the denominator is that, with 1000 retrievals, it is impossible to retrieve all the 1300 relevant documents, so we divide by the most possible instead of all the relevant documents. With this metric, method 1 now has an AP@5 of 0.2, and method 2 has an AP@ 5 of 0.42 which reflects the reality of the retrieval. The only change needed is the replacement of ``relevant\_num'' in this line\footnote{\url{https://github.com/thuml/HashNet/blob/master/caffe/models/predict/imagenet/predict_parallel.py\#L42}}
with ``R'' for the ImageNet-100 setting.

Table~\ref{tab:hashing_error} presents a comparison of both MAP@1000 computed as reported, RMAP and corrected, CMAP for \llc and HashNet. This shows a stark drop in the MAP values along with a non-intuitive correlation between RMAP and CMAP making off-the-shelf comparisons harder. Lastly, the MAP computation from HashNet also skipped the AP values when none of the relevant documents were retrieved (which is extremely rare at 1000 retrieval samples) but the AP score should have been 0 for that particular case. We fixed that in both RMAP and CMAP metrics reported here.

\section{Hyperparameters}
\label{sec:hparams}

\paragraph{Classification}
The first phase of the classification pipeline, ie., codebook learning, uses the hyperparameters - SGD+Momentum optimizer, batch size of 256, cosine learning rate routine, and 100 epochs - used in training the standard 1000-way linear classifier using a ResNet50~\citep{wortsman2019discovering,kusupati2020soft}. However, given the warm-starting of the second phase using the learnt backbone and the availability of the learnt codebook, the training routine of the second phase runs for about 25 epochs with a reduced learning rate of 0.01.

\paragraph{Retrieval}

As we start with pre-trained models for retrieval. We use the same hyper-parameters as the second phase of the classification pipeline except for the learning rate. We use a much learning rate of 0.01 for AlexNet and 0.1 for ResNet50. The second phase of learning instance codes uses a reduced learning rate of 0.005 and 0.01 for AlexNet and ResNet50 respectively. The rest of the hyperparameters and training mechanisms are the same as the standard ResNet50 training.

\section{Quantitative Evaluation of Hierarchy \& More Visualizations}
\label{sec:quant_hier}

Quantitative evaluation of the discovered hierarchies is a hard problem because we can not directly compare ours to the original ImageNet hierarchy which is not binary unlike ours. However, one proxy way is to use the pair-wise inner-product heat map (see Figure~\ref{fig:heat}) of 2048 dimensional representation as the base and compare using row-wise (class-wise) ranking metrics like Spearman's rank correlation coefficient in which the learnt 20-bit codebook has a mean coefficient across all class of 0.3 compared to 0.005 of a random codebook. Lastly, the number of unique class codes (ideally should be equal to $L$) learnt as part of the codebook plays an important role as the code collapse leads to loss of information about multiple classes resulting in unclear decoding when required. 

\begin{figure}[b]
\centering
\resizebox{\columnwidth}{!}{%
\includegraphics[width=0.5\columnwidth, angle =-90 ]{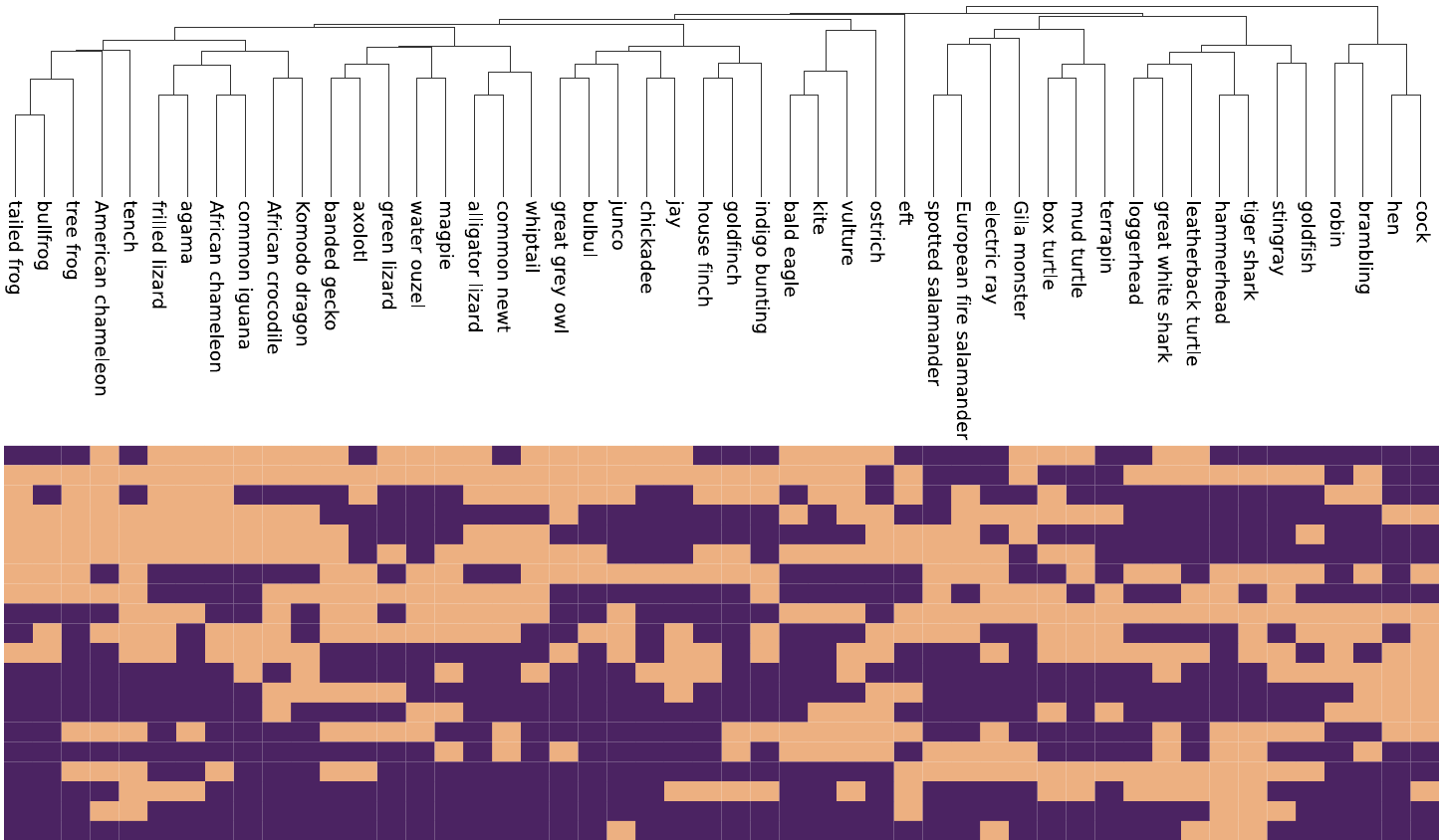}
}
\caption{\small Discovered hierarchy on 50 classes of ImageNet-1K along with the corresponding class codes. Purple is $+1$ and beige is $-1$ in each class code corresponding to the species in the dendrogram.}
\label{fig:heir_code}
\end{figure}
\begin{figure}[b!]
\centering
\resizebox{\columnwidth}{!}{%
\begin{tabular}{cc}
  \includegraphics[width=.5\textwidth]{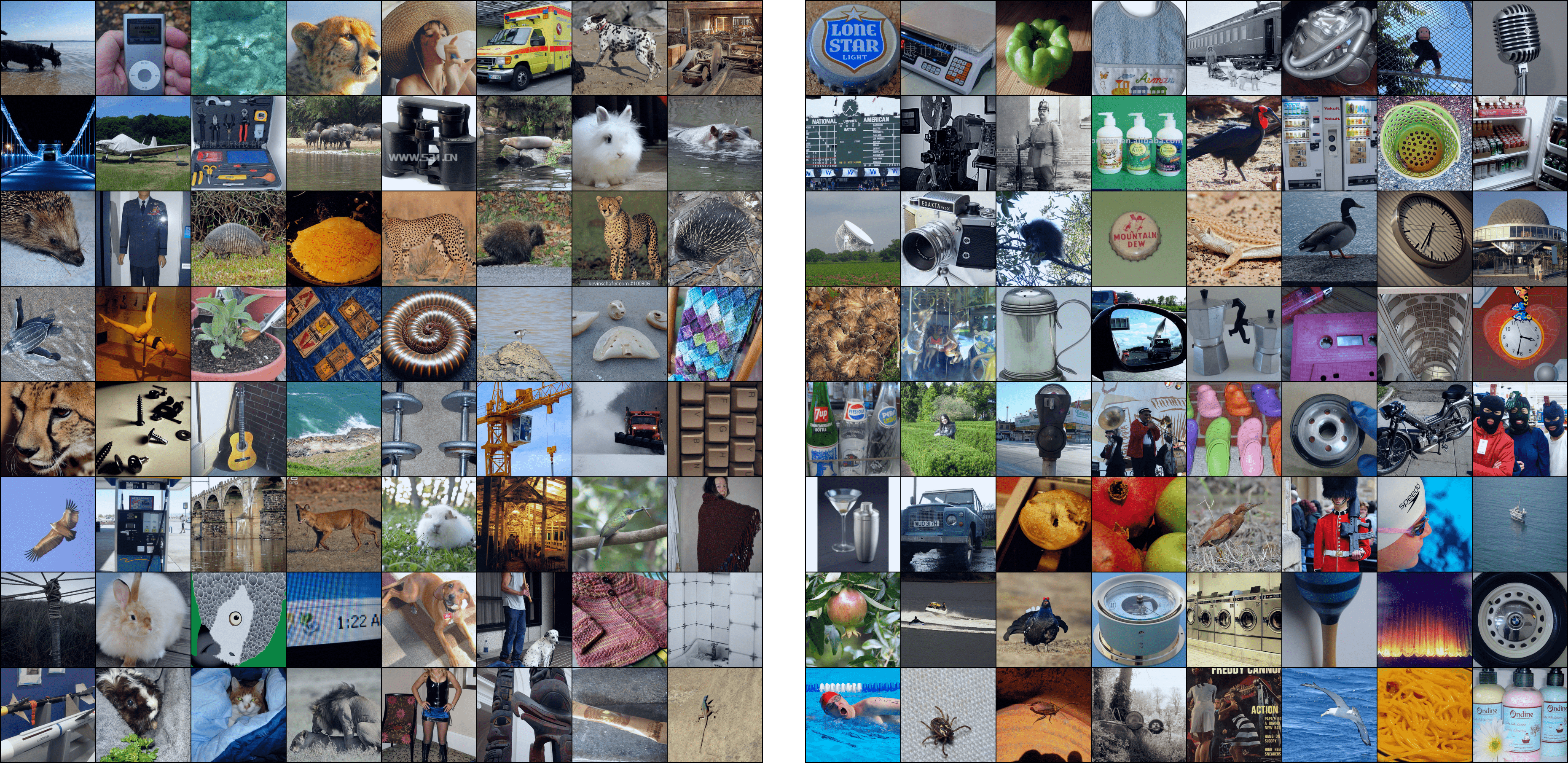}&
  \includegraphics[width=.5\textwidth]{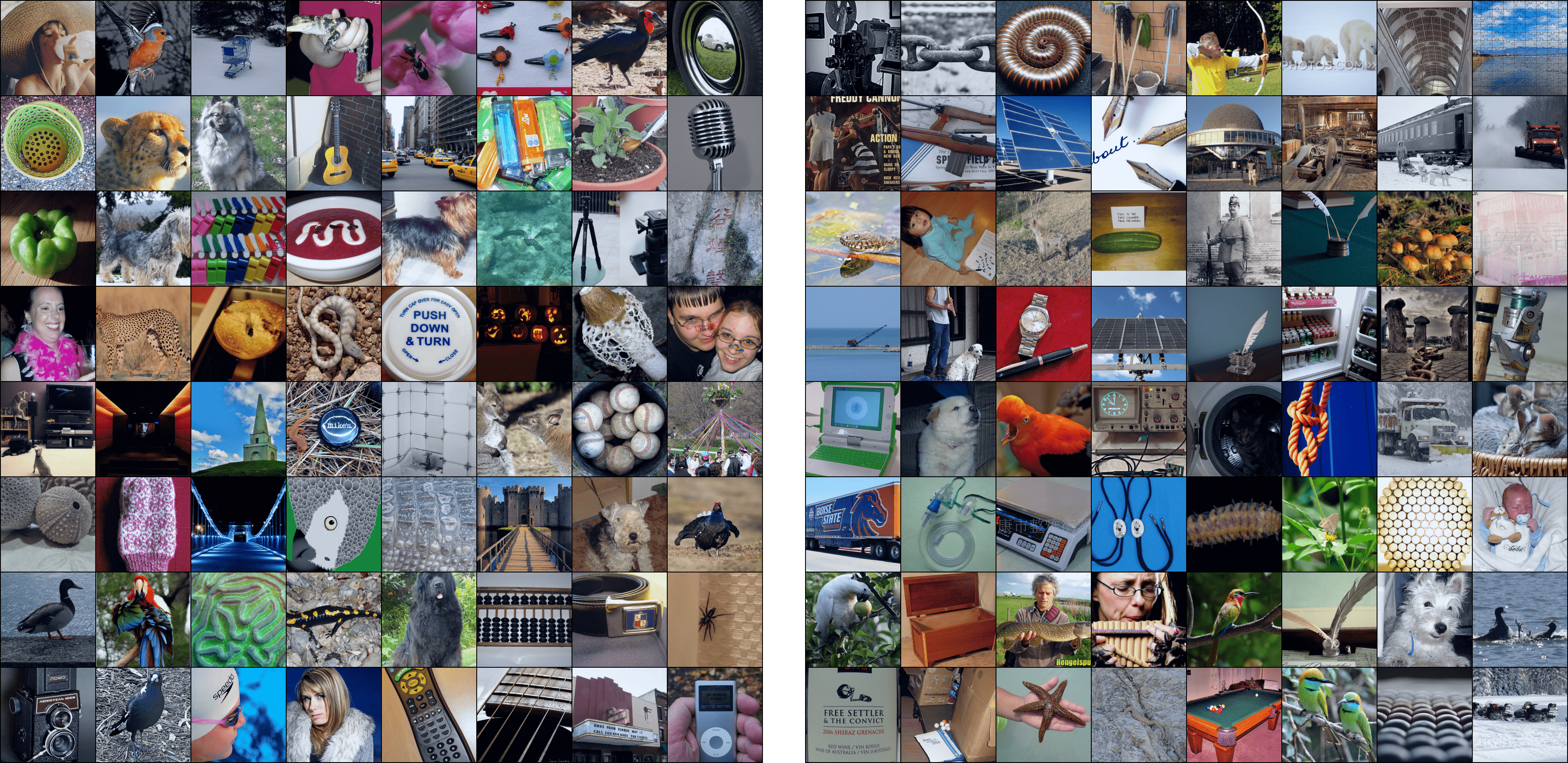}\\
(a)&(b)
\end{tabular}
}
\caption{\small The hyperplanes of bit numbers (a) $0$ \& (b) $16$ visualized using the images being split on either side $+1, -1$ (left and right of the white bar in each of (a) and (b)), and are sorted using the probability. These splits do not show any trivial attributes being discovered, but often we find bits which do fine grained classification between close classes.}

\label{fig:bit-splits}
\vspace{-2mm}
\end{figure}

It would be exciting to compare the discovered taxonomy to the original WordNet~\citep{miller1995wordnet} hierarchy based on which the ImageNet-1K was curated. However, the major roadblock comes when we realize that our discovered hierarchy is binary while WordNet is $k$-ary making fair comparison almost impossible. We also explored the idea of cost-sensitive metrics~\citep{deng2011hierarchical} based on hierarchy to evaluate the classification but fell short due to the same limitation of unfairness in comparing binary to $k$-ary.

Lastly, we also wanted to see if the learnt bit codes/hyperplanes result in splitting the images by discovering (potentially interpretable) attributes as with previous works~\citep{rastegari2012attribute}. However, unlike previous works, which use simple features for encoding images, we learn a highly non-linear representation using deep neural networks. This resulted in learnt bit code hyperplanes that often split the images using highly non-linear, non-trivial, and non-interpretable attributes. Figure~\ref{fig:bit-splits} shows the images on either side of the hyperplane of the corresponding bit sorted by the confidence of being $+1$ or  $-1$ (probability from the logistic function for binary classification). While, with deeper analysis and further visualization, we might be able to deduce what is being learnt, but at the surface level, without explicit learning with priors, the discovered attributes are non-interpretable affecting zero-shot and few-shot capabilities of \llc based models.

\section{Out-of-Distribution (OOD) Detection Experiments}
\label{sec:ood_expts_app}

OOD detection for a multi-class classification model can be achieved by simple baselines that set a threshold on a heuristic based on the prediction probabilities~\citep{hendrycks2016baseline}. For the classification models trained with ResNet50 on ImageNet-1K, we evaluate on two datasets which we consider as out-of-distribution to ImageNet-1K that has 50K samples as the in-distribution validation set. Our method uses a 20-bit classification model while the baselines use the pre-trained ResNet50 with linear classifier. 

ImageNet-750~\citep{wallingford2020overfitting} is a long-tailed dataset of 750 classes with about 69K samples. These 750 classes were sampled from ImageNet-22K~\citep{deng2009imagenet} ensuring no clash with the ImageNet-1K subset. All the instances in this dataset are OOD to a model trained on ImageNet-1K. MIT Places~\citep{zhou2014learning} dataset is aimed at scene recognition rather than object recognition, unlike ImageNet-1K. All the $\sim$18K samples in the validation set of Places365 were also treated as OOD to ImageNet-1K during this evaluation.

\begin{figure}[ht!]
\centering\hspace*{-3ex}
\begin{tabular}{cc}
  \includegraphics[width=.4\textwidth]{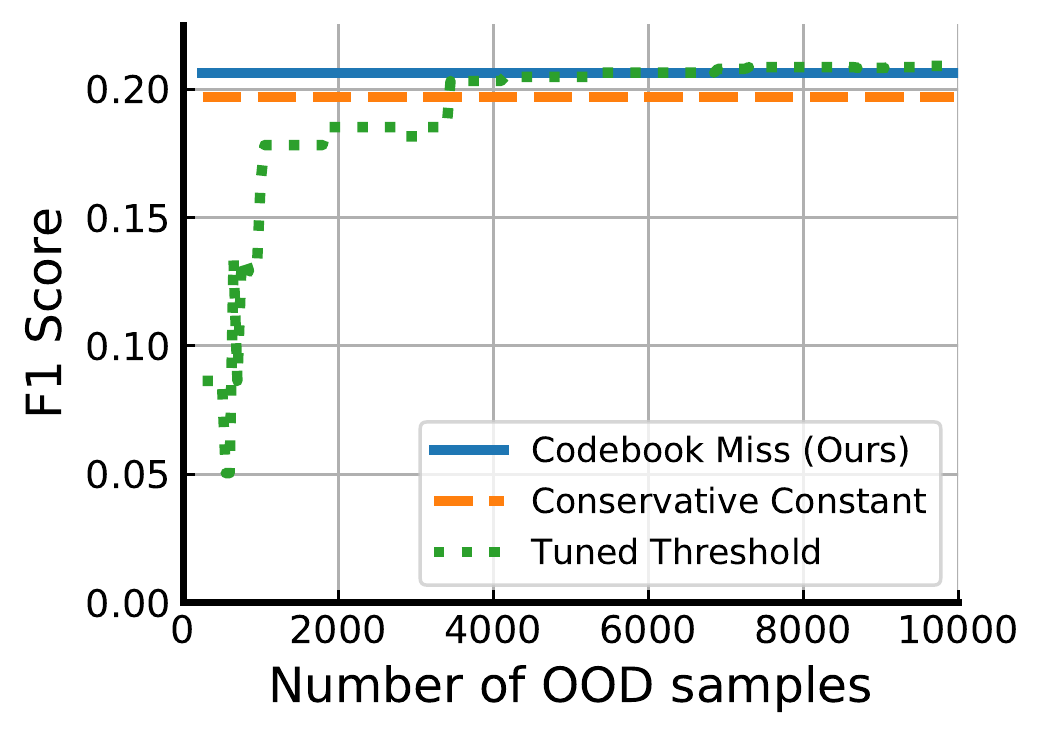}&
  \includegraphics[width=.4\textwidth]{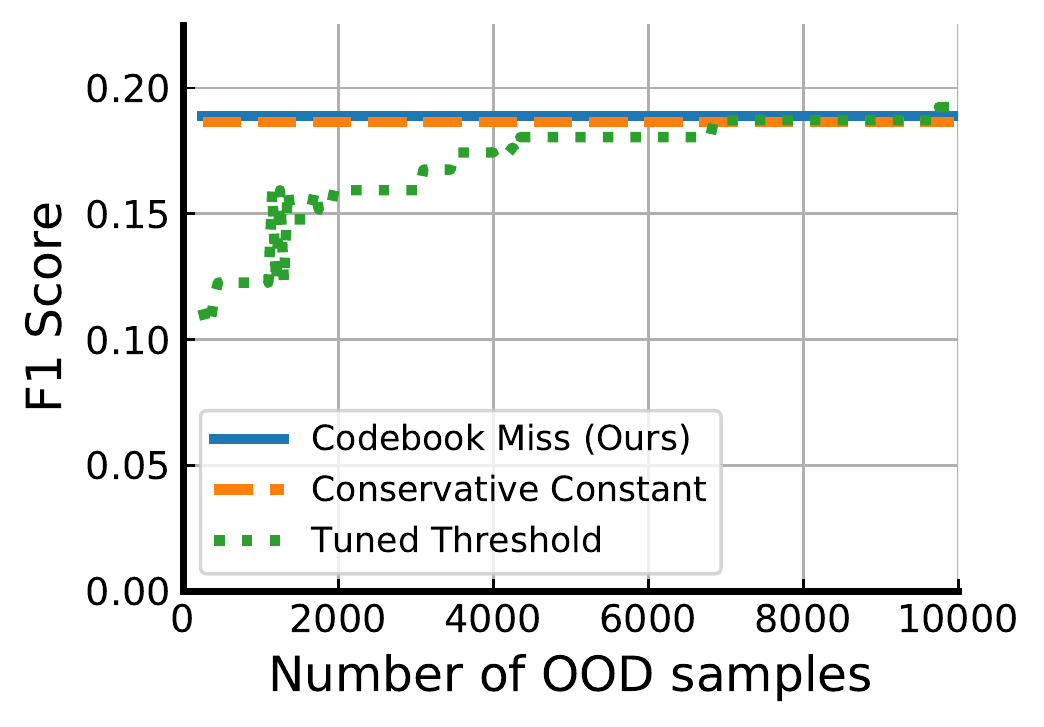}\\
(a)&(b)
\end{tabular}
\vspace{-1mm}
\caption{\small OOD detection performance when the validation set is not representative of the deployment setting for a) ImageNet-750 \& b) MIT Places datasets evealuted on a ResNet50 model trained for ImageNet-1K.}
\label{fig:OOD_noratio}
\vspace{-2mm}
\end{figure}

As mentioned in Section~\ref{sec:ood_expts}, \llc based OOD detection looks at the encoded instance code and classifies a sample as OOD in case of a missing exact match in the learnt codebook. This does not require any OOD samples whatsoever apriori to the deployment. We consider two baselines: a) tuning of the threshold on the probability of maximum logit to maximize F1 score using a validation set with OOD examples~\citep{hendrycks2016baseline}, and b) setting a conservative threshold on the probability of maximum logit which is 1 standard deviation greater than the mean probability of the maximum logit using just 50 OOD examples. Note that the performance of the first baseline varies with i) the ratio of in distribution to OOD samples in the validation set that is being used for tuning the threshold, ii) the setup with a different validation to test out-of-distribution ratios.

Our primary focus is on the setting where the validation set used for tuning thresholds is not representative of the test set. In this setting the in distribution samples are constant, all the 50K samples. We only vary the number of OOD samples while tuning the threshold. The final OOD performance using the F1 score was measured on a test set with all the 50K in distribution samples along with a random 10K samples from the OOD set (ImageNet-750 \& MIT Places).

Figure~\ref{fig:OOD_noratio} captures the effectiveness of \llc based OOD detection in the setting where validation is not representative of test. Our out-of-the-box OOD detection method is at least as effective as the baselines. Remarkably, the tuning of the threshold to maximize F1 using a validation set requires at least 3000 OOD samples to even get close to our method which requires {\em no samples}. 

The second setting is where the validation and test sets have the same number of OOD samples for a fixed in-distribution set. Here the validation and test phases have same number of re-sampled OOD instances. All the 50K in distribution samples are used for both validation and testing.

\begin{figure}[ht!]
\centering

\begin{tabular}{cc}
  \includegraphics[width=.4\textwidth]{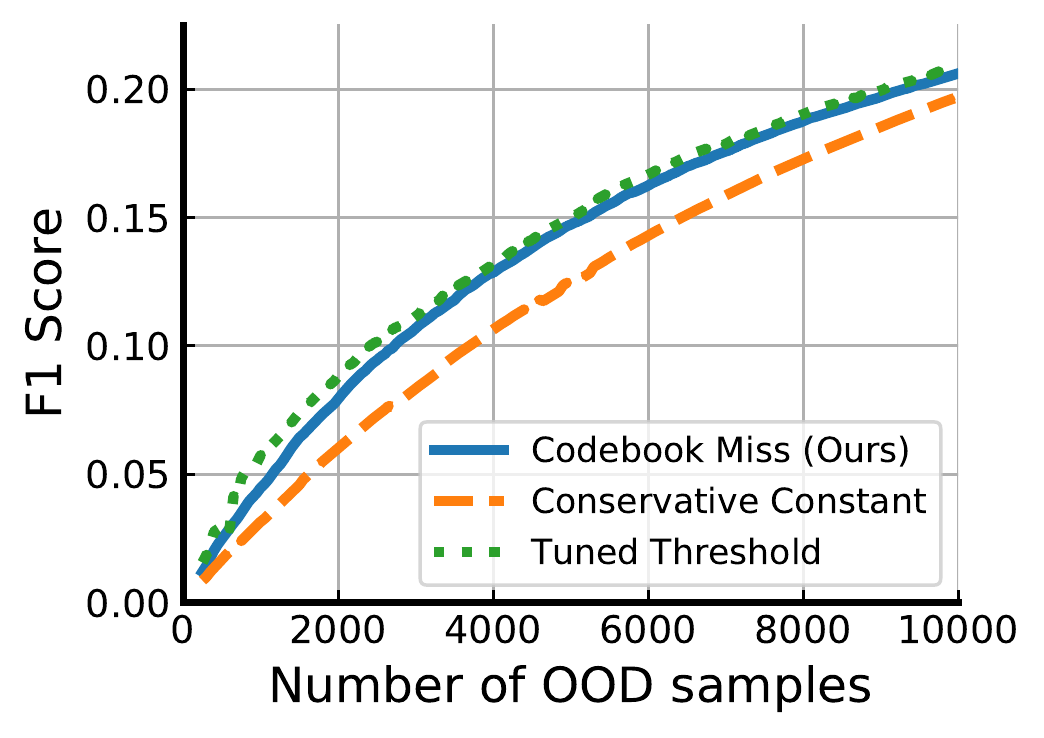}&
  \includegraphics[width=.4\textwidth]{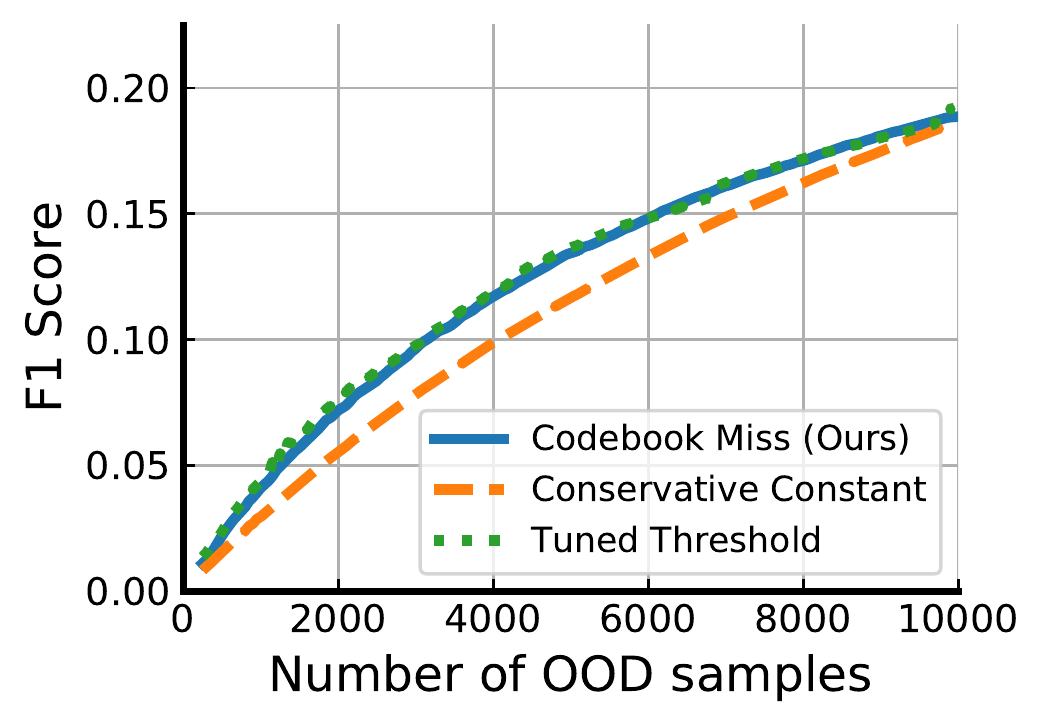}\\
(a)&(b)
\end{tabular}
\vspace{-1mm}
\caption{\small OOD detection performance when the validation set is representative of the deployment setting for a) ImageNet-750 \& b) MIT Places datasets evaluated on a ResNet50 model trained for ImageNet-1K.}
\label{fig:OOD_ratio}
\end{figure}

Figure~\ref{fig:OOD_ratio} shows that \llc based OOD detection is as competitive as the tuned threshold baseline~\citep{hendrycks2016baseline} which is optimizing F1 for the exact same ratios. These experiments help in arguing that \llc based classification models have a strong, sample-efficient inherent OOD detection capability. which can be used for sequential learning, to detect and continuously add new classes, along with robustness.

\end{document}